\documentclass[preprint]{elsarticle} %for archive
\usepackage{lineno,hyperref}
\usepackage{multicol}
\usepackage{multirow}
\usepackage{graphicx}
\usepackage[table,xcdraw]{xcolor}
\usepackage{caption}
\usepackage{enumitem}
\usepackage{soul}
\usepackage{hyperref}
\modulolinenumbers[5]
\usepackage[normalem]{ulem}
\usepackage{dsfont}
\usepackage[ruled,vlined]{algorithm2e}
\usepackage{algorithmicx} % <===========================================
\usepackage{algpseudocode} % <==========================================
\usepackage[bb=boondox]{mathalfa}

\journal{Journal of Computer Speech and Language}

\begin{document}
\begin{frontmatter}
% \title{What do End-to-End Speech Models Learn about Speaker, Language and Channel Information? \\\\\large{A Layer-wise and Neuron-level Analysis}}
\title{What do End-to-End Speech Models Learn about Speaker, Language and Channel Information? \leavevmode\\\large{A Layer-wise and Neuron-level Analysis}}

\author{Shammur~Absar~Chowdhury\corref{cor}} 
\ead{\{shchowdhury\}@hbku.edu.qa}
\cortext[cor]{Corresponding author}
\author{Nadir~Durrani}
\ead{\{ndurrani\}@hbku.edu.qa}
\author{Ahmed~Ali}
\ead{\{amali\}@hbku.edu.qa}
\address {Qatar Computing Research Institute, \\HBKU, Doha, Qatar}

\fntext[myfootnote]{\copyright~2017. This manuscript version is made available under the CC-BY-NC-ND 4.0 license http://creativecommons.org/licenses/by-nc-nd/4.0/
}
% \maketitle
% \end{frontmatter}
\begin{abstract}

Deep neural networks are inherently opaque and challenging to interpret. Unlike hand-crafted feature-based models, we struggle to comprehend the concepts learned and how they interact within these models. This understanding is crucial not only for debugging purposes but also for ensuring fairness in ethical decision-making. In our study, we conduct a post-hoc functional interpretability analysis of pretrained speech models using the probing framework \cite{hupkes2018visualisation}.
Specifically, we analyze utterance-level representations of speech models trained for various tasks such as speaker recognition and dialect identification. We conduct layer and neuron-wise analyses, probing for speaker, language, and channel properties. Our study aims to answer the following questions: i) what information is captured within the representations? ii) how is it represented and distributed? and iii) can we identify a minimal subset of the network that possesses this information?  
 Our results reveal several novel findings, including: i) channel and gender information are distributed across the network, ii) the information is redundantly available in neurons with respect to a task, iii) complex properties such as dialectal information are encoded only in the task-oriented pretrained network, iv) and is localised in the upper layers, v) we can extract a minimal subset of neurons encoding the pre-defined property, vi) salient neurons are sometimes shared between properties, vii) our analysis highlights the presence of biases (for example gender) in the network. Our cross-architectural comparison indicates that: i) the pretrained models capture speaker-invariant information, and ii) CNN models are competitive with Transformer models in encoding various understudied properties.

\end{abstract}
\begin{keyword} \texttt{Speech} \sep \texttt{Neuron-level Analysis} \sep \texttt{Interpretibility} \sep \texttt{Diagnostic Classifier} \sep \texttt{AI explainability} \sep \texttt{End-to-End Architecture} \end{keyword}
\end{frontmatter}
% 

% \linenumbers
\section{Introduction}
\label{sec:intro}

Deep Neural Networks (DNNs) have constantly pushed the state-of-the-art in all arenas of Artificial Intelligence including Natural Language Processing (NLP) \cite{deng2018deep} and Computer Vision (CV) \cite{voulodimos2018deep}. Impressive strides have also been made in speech technologies, for example, Automatic Speech Recognition (ASR) \cite{amodei2016deep,miao2015eesen,pratap2019wav2letter++,chan2016listen,chowdhury2021onemodel,ali2021csmodel}, Pretrained Speech Transformers \cite{baevski2020wav2vec,liu2020mockingjay,liu2020tera,chi2020audio}, Dialect, Language and Speaker Identification \cite{jin2017end, trong2016deep, heigold2016end, nagrani2017voxceleb, snyder2017deep, shon2018convolutional, snyder2018x} models. While deep neural networks provide a simple and elegant end-to-end framework with a flexible training mechanism, the output models are inherently black-box. In contrast to traditional models, they lack an explanation of what knowledge is captured within the learned representations, and how it is 
used by the model during prediction. The opaqueness of deep neural models hinders practitioners from understanding their internal mechanics, which is crucial not only for debugging but also for ethical decision-making and fairness in these systems \cite{doshi2017towards,lipton2018mythos}. These internal mechanics include metrics that are often as important as the models' performance. As a result, a plethora of research has been conducted to investigate how deep neural models encode auxiliary knowledge in their learned representations through classifiers \cite{dalvi2019one, durrani2022linguistic, belinkov2017neural, dalvi2017understanding}, visualizations \cite{zhang2018visual, zeiler2014visualizing}, ablation studies \cite{sheikholeslami2021autoablation,bau:2019:ICLR}, and unsupervised methods \cite{harwath2017learning, dalvi2022discovering,durrani-etal-2022-latent}.

In this paper, we focus our efforts on studying pretrained speech models. It is challenging to understand and interpret the outputs of speech models, mainly due to the variable lengths of the input signals and the complex hierarchical structures that exist at different time scales.\footnote{For instance, syllables are made up of phonemes that temporally extend over a few milliseconds. These syllables are then grouped into higher orders such as words, intonational/prosodic phrases, and sentences, which range from a few hundred milliseconds to seconds.}
% In this paper, we focus our efforts on studying pretrained speech models. Understanding and interpreting speech model outputs are challenging, mainly due to the variable lengths of the input signals, and the complex hierarchical structures that exist at different time scales.\footnote{For instance, syllables are made up of phonemes that temporally extend over a few milliseconds, which are then grouped into higher orders like words, intonational/prosodic phrases, and sentences ranging from a few hundred milliseconds to seconds.} 
% and exhibit complex hierarchical structures on different time scales.\footnote{For instance, syllables are made up of phonemes that temporally extend over a few milliseconds, which are then grouped into higher orders like words, intonational/prosodic phrases, and sentences ranging from a few hundred milliseconds to seconds. } 
Furthermore, environmental factors such as channel information (e.g., signal recording and transmission quality of speech) and speaker information (voice identity, gender, age, and language) can influence the characteristics of the input signal; and make it more challenging to comprehend the models' decisions.
% influence the characteristics of the input signal, and make it more difficult to comprehend the models' decisions.
% Moreover, the environmental (e.g. \textit{channel information} -- signal recording and transmission quality of the speech) and speaker (\textit{voice identity, gender, age, language}) influenced variables 
% in speech models to impose different characteristics on the acoustic input, resulting in added complexity, making 
% and make it harder to understand models' decisions. 
% For example, in speaker recognition (SR) systems, the model process personal data,\footnote{carrying information that identifies an individual,  for e.g., speaker's identity or ethnic origin from his/her voice} and is highly susceptible to biases \cite{6613025}, therefore it is fundamental that we study and interpret what other factors like gender, race, accent, microphones are hindering the model's decision. While some work has been done to interpret representations in the speech models \cite{beguvs2021interpreting, elloumi2018analyzing, chowdhury2020does,shah2021all,wang2017does,becker2018interpreting}, no prior work has been carried out to do a fine-grained neuron-level analysis.
Speaker recognition (SR) systems, for instance, involve the processing of personal data,\footnote{carrying information that identifies an individual, e.g., speaker's identity or ethnic origin from his/her voice} and are highly susceptible to biases \cite{6613025}. Therefore, it is essential to study and interpret other factors that could influence the model's decision, such as gender, race, accent, and the characteristics of the microphone used. While some prior research has focused on interpreting the representations in speech models \cite{beguvs2021interpreting, elloumi2018analyzing, chowdhury2020does,shah2021all,wang2017does,becker2018interpreting}, no fine-grained neuron-level analysis has been carried out.

A prominent framework for probing in the NLP domain is the Probing Classifiers \cite{belinkov2017neural}, which has been extensively used to analyze representations at the level of individual layers \cite{elloumi2018analyzing, chowdhury2020does, shah2021all, wang2017does}, attention heads \cite{ghader2017does}, and a more fine-grained neuron level \cite{dalvi2019one, durrani2020analyzing, qian2016analyzing}. These studies reveal interesting findings, such as how different linguistic properties, such as morphology and syntax, are captured in different parts of the network, and how certain properties are more localized or distributed than others. Furthermore, the Probing Classifiers framework has been applied to the analysis of individual neurons, enabling a more comprehensive understanding of the network \cite{girshick2016, nguyen2016synthesizing, bau2017network, dalvi2019one, durrani2020analyzing, qian2016analyzing, shi2016neural}. It also holds significant potential for various applications, including system manipulation \cite{bau:2019:ICLR} and model distillation \cite{rethmeier2019txray, frankle2018lottery}.
% Moreover, the paradigm has been used to analyze individual neurons, facilitating a deeper understanding of the network \cite{girshick2016,nguyen2016synthesizing, bau2017network,dalvi2019one,durrani2020analyzing, qian2016analyzing,shi2016neural}, and entails many potential applications such as system manipulation \cite{bau:2019:ICLR} or model distillation \cite{rethmeier2019txray,frankle2018lottery}.
%benefits such as manipulating system's output \cite{indivdualneuron:arxiv19} while debasing the network w.r.t certain property (like gender or racial elements), model distillation and compression \cite{rethmeier2019txray,frankle2018lottery}, domain adaptation \cite{gu2021pruningthenexpanding},  feature selection for downstream tasks \cite{dalvi2020analyzing} and guiding architectural search etc.

%In this paper 
% We use probing classifiers to carry out a post-hoc functional interpretation of pretrained speech models. We carry out a layer-wise and fine-grained neuron-level analysis on the pretrained speech models for i) speaker: gender and voice identity, ii) language and its dialectal variants, and iii) channel information, using utterance-level representation. We carry out our study with the following research questions:
% i) does the end-to-end speech model capture different properties (e.g. gender, channel or speaker information)? ii) where in the network is it represented and how localized or distributed is it?

We use probing classifiers to conduct a post-hoc functional interpretation. %of pretrained speech models. 
Our analysis includes a layer-wise and fine-grained neuron-level examination of the pretrained speech models, specifically focusing on: i) speaker information such as gender and voice identity, ii) language and its dialectal variants, and iii) channel information, using utterance-level representation. Our study is guided by the following research questions: i) does the end-to-end speech model capture different properties (e.g. gender, channel, or speaker information)? ii) where in the network is it represented, and how localized or distributed is it?

% #how we want to achieve it
% We use aforementioned framework based on probing classifiers \cite{hupkes2018visualisation,conneau2018you}. 
We follow a simple methodology: i) extract the representations (of speech utterances) from our studied pretrained model; ii) train an auxiliary classifier towards predicting the studied properties; iii) the accuracy of the classifier serves as a proxy to the quality of the learned representation with respect to that property. We train classifiers using different layers to carry out our layer-wise analysis and do a fine-grained analysis by probing for neurons that capture these properties.

% #what we used
% We investigate 4 pretrained end-to-end architectures: 2 Convolutional Neural Networks (CNN) architectures trained towards the task of  \textbf{\emph{i) speaker recognition}} and  \textbf{ii) \emph{dialect identification}}, 
% and 2 Transformer architectures trained to \textbf{\emph{iii) reconstruct the masked signal}}.\footnote{Please refer to \cite{WangSpeech} for pre-training speech representations via a masked reconstruction loss.} The choice of architectures was driven by the fact that these models have shown state-of-the-art performances in different speech tasks. Also comparing CNNs to the recently emerged transformer architecture provides interesting cross-architectural analysis. We train our probes towards the following extrinsic properties of interest:  \textbf{\emph{ i) gender classification}},  \textbf{\emph{ ii) speaker verification}}, \textbf{\emph{ iii) language identification}}, \textbf{\emph{iv) dialect}}, and \textbf{\emph{v) channel classification}}.
We investigate four pretrained end-to-end architectures: two Convolutional Neural Networks (CNN) architectures trained for the tasks of \textbf{\emph{i) speaker recognition}} and \textbf{ii) \emph{dialect identification}}, as well as two Transformer architectures trained to \textbf{\emph{iii) reconstruct the masked signal}}.\footnote{Please refer to \cite{WangSpeech} for information on pre-training speech representations using masked reconstruction loss.} We chose these architectures because they have demonstrated state-of-the-art performance in various speech tasks. Additionally, comparing CNNs to the recently emerged Transformer architecture provides interesting cross-architectural analysis. Our probes are trained to capture the following extrinsic properties of interest: \textbf{\emph{i) gender classification}}, \textbf{\emph{ii) speaker verification}}, \textbf{\emph{iii) language identification}}, \textbf{\emph{iv) dialect}}, and \textbf{\emph{v) channel classification}}. Our cross-architectural analysis shows that:

 \begin{itemize}
 \setlength\itemsep{0em}
    
    \item \textit{Minimal Neuron Subset:} 
    %Few neurons ($\approx$ 1-5\%) are required to encode simple properties (such as gender)
    Simple properties, such as gender, can be encoded with minimal neuron activation (approximately 1-5\%).\footnote{We consider gender as a simple property due to its high modeling performance, with accuracy exceeding 90\% \cite{levitan2016automatic}. This performance is achieved using well-established distinguishing features such as intonation, speech rate, duration, and pitch, among others, with classical rule-based/machine learning techniques} Complex properties such as regional dialect identification\footnote{Discriminating between fine-grained acoustic differences in dialects of the same language family with shared phonetic and morphological inventory poses a significant challenge. The model's ability to do so directly affects its performance, as highlighted by \cite{chowdhury2020does}.} or voice identification requires a larger subset of neurons to achieve optimal accuracy. %($\rightarrow$ RQ1, RQ3)
    
    \item \textit{Task-specific Redundancy:} 
    Gender and channel information are redundantly observed in both layer and neuron analyses. These phenomena are distributed across the network and can be captured redundantly with a small number of neurons from any part of the network. Similar redundancy is also observed in encoding language properties.
    
    \item \textit{Localized vs Distributive:} The salient neurons are localized in the final layers of the pretrained models, indicating that the deeper layers of the network are more informative and capture more abstract information. This observation  
    was true for both simple and complex tasks. 
    
    \item \textit{Robustness:} Most of the understudied pretrained models learn speaker-invariant information. 
    
    \item \textit{Polysemous Neurons:} Neurons are multivariate in nature and shared among properties. For example, some neurons were found to be salient for both voice identity and gender properties. 
    
    \item \textit{Bias:}  
    We were able to highlight neurons responsible for sensitive information (e.g., gender). This can be potentially  
    useful to mitigate representation bias in the models. 
    
    \item \textit{Pretrained Models for Transfer learning:} Pretrained convolutional neural networks (CNNs) can provide comparable, or even superior, performance compared to speech transformers. These CNNs are commonly used as feature extractors or for fine-tuning in downstream tasks. In line with the findings of \cite{tay2021pretrained}, our results highlight the potential of utilizing large pretrained CNNs as effective alternatives to pretrained speech transformers.
    
    % Pretrained CNNs give comparable performance, sometimes even outperform the speech transformers. %and %can %easily 
    % %be 
    % These CNNs are used as a viable feature extractor (or for fine-tuning) in the downstream tasks. In line with \cite{tay2021pretrained}, our findings suggest the potential of re-using large CNNs as pretrained networks with similar efficacy as the pretrained speech transformers. 
    
    \end{itemize}

\noindent %To the best of our knowledge, our work is the first attempt that carries large-scale neuron analysis. 
To the best of our knowledge, our study represents the first attempt at conducting a large-scale analysis of neurons.  While other contemporary works, such as
%Other contemporary works, for e.g.,
\cite{chung2021similarity, chowdhury2020does, wang2017does,yang2022autoregressive,beguvs2021interpreting,shah2021all,pasad2022comparative} have largely focused on layer-wise analysis, our work goes beyond and delves into a more fine-grained, neuron-level analysis. This microscopic view of speech representation opens up potential applications, including neuron manipulation and network pruning, among others, which can greatly benefit from our findings.

\section{Related Work}
\label{sec:related}

The rise of deep Neural Networks has seen a subsequent rise in interpretability studies, due to the black-box nature of these models. One of the commonly used interpretation technique is the probing-tasks or the diagnostic-classifiers framework \cite{belinkov2017neural}. 
This approach has been used to probe for different linguistic properties captured within the network. For example researchers probed for: i) morphology using attention weights \cite{ghader2017does}, or recurrent neural network (RNN)/transformer representations \cite{peters2018dissecting,shi2016does,blevins2018deep}, in neural machine translation (NMT) \cite{belinkov2017neural,dalvi2017understanding} and language models (LM) neurons \cite{dalvi2019one, dalvi2019neurox},
ii) anaphora \cite{voita2018context}, iii) lexical semantics with LM and NMT states \cite{belinkov2017neural,dalvi2017understanding}, and iv) word presence \cite{liu2018lstms}, subject-verb-agreement \cite{linzen2016assessing}, relative islands \cite{chowdhury2018rnn}, number agreement \cite{gulordava2018colorless}, semantic roles \cite{ettinger2016probing}, syntactic information \cite{shi2016does,linzen2016assessing,conneau2018you,chowdhury2018rnn,merlo2019probing} among others using hidden states. Detailed comprehensive surveys are presented in \cite{belinkov2019analysis,sajjad-etal-2022-neuron}. 

In the arena of speech modeling, a handful of properties have been examined, namely: i) utterance length, word presence, and homonym disambiguation using audio-visual RNN states \cite{chrupala2017representations}; ii) phonemes and other phonetic features using CNN activation in end-to-end speech recognition, \cite{nagamine2015exploring,nagamine2016role, chaabouni2017learning} using activations in the feed-forward network of the acoustic model, \cite{silfverberg2021rnn} using RNNs; along with iii) formants and other prosodic features examined in the CNNs trained from raw speech \cite{beguvs2021interpreting}; iv) gender \cite{nagamine2015exploring, wang2017does, chowdhury2020does}; v) speaker information \cite{wang2017does,chowdhury2020does}, style, and accent \cite{elloumi2018analyzing} using network activations; vi) channel \cite{wang2017does, chowdhury2020does} using activations; vii) fluency, pronunciation, and other audio features from transformers \cite{shah2021all}; viii) word properties, phonetic information using several self-supervised model layers \cite{pasad2022comparative}; ix) frame classification, segment classification, fundamental frequency tracking, and duration prediction using the autoregressive predictive coding (APC) approach \cite{yang2022autoregressive}; and x) linguistic information from emotion recognition transformer models \cite{triantafyllopoulos2022probing}. Apart from classification, other methods for finding associations include:

\begin{itemize}
\setlength\itemsep{-0.3em}
    \item Computing correlations or similarities: e.g. with other acoustic features \cite{wu2016investigating}, or similarity between representations learned by different models \cite{chung2021similarity}
    \item Regression: e.g. sentence length with encoder neurons in NeuralMT \cite{shi2016neural}
    \item Clustering: e.g. word class using A/V CNN embeddings \cite{harwath2017learning}
    \item Detecting change point of activation: e.g. using RNN gates for finding phoneme boundaries \cite{wang2017gate} 
    \item Visualisation: e.g. in deep CNN models by sampling image patches \cite{girshick2016,zhou6856object,bauvisualizing} or by generating images \cite{nguyen2016synthesizing} that maximize the activation of each individual neuron among others
\end{itemize}

% Our study draws motivation from and 
Building upon the research conducted in \cite{dalvi2019one,chowdhury2020does}, our approach leverages proxy classifiers to gain valuable insights into the encoded information within pretrained deep learning networks. However, our approach differs from theirs in that we have analyzed various types of pretrained models, incorporating different objective functions and architectures. In addition to analyzing the dialectal model, ADI, as utilized in \cite{chowdhury2020does}, we have further explored a speaker recognition model with a similar architecture. %Moreover,
% Furthermore, is one of the first to probe speech transformers, trained using self-supervision. 
% \textit{Moreover, our study is the first to probe pretrained speech models to conduct fine-grained neuron-level analysis.} 
\textit{Furthermore, our research represents the pioneering effort in exploring pretrained speech models through fine-grained neuron-level analysis.}

Diverging from the approaches taken in \cite{wang2017does, chowdhury2020does}, our study focuses on analyzing individual or groups of neurons that effectively encode specific properties. Taking inspiration from \cite{dalvi2019one}, which probed neurons learning linguistic properties in deep language models, our research represents the first attempt to closely examine the neurons within pretrained speech networks that capture speaker, language, and channel properties.

% \todo{Repeated many times: this is the first attempt...}

%TODO later 
% \nd{May be also touch upon some literature on causation analysis and amnesic probing. Not super important and can be done in camera-ready also I guess}
\section{Methodology}
\label{sec:methods}

Our methodology is based on the probing framework called the \emph{Diagnostic Classifiers}. 
We extract activations from the pretrained neural network model $\mathds{M}$ and use these as static features to train a classifier $\mathds{P}$ towards the task of predicting a certain property.
%We train a  classifier using the activations generated from the pretrained neural network $\mathds{M}$, with $L$ layers: $\{l_1, l_2, \ldots, l_L\}$, as static features, towards the task of predicting a certain property (See Figure \ref{fig:pipeline}). 
The underlying assumption is that if the classifier $\mathds{P}$ successfully predicts the property, the representations implicitly encode this information. 
%We train layer- and neuron-wise probes using logistic-regression classifiers. More details is in Section \ref{ssec:ul-representation} and Section \ref{ssec:proxy_classifier}.
% Formally, consider a pre-trained neural speech model $\mathds{M}$ with $L$ layers: $\{l_1, l_2, \ldots, l_L\}$. Given a dataset of $N$ utterances, $D=\{u_1, u_2, ..., u_N\},$ consisting of varying acoustic feature frames, we first use a pooling function ($\varphi^{[l_i]}$, $l_i \in L$), to extract the utterance level representation $U_n$ ($=\varphi^{[l_i]}(u_n)$).
% % , using a pooling function ($\varphi^{[l_i]}$). 
% This latent representation corresponds with a class ($c$) label from a set of annotations $C=\{c_1, c_2, ..., c_C\}$.
% % , we map each word $s_i$ in the data $\sD$ to a sequence of latent representations: $\sD\xmapsto{\modelM}\zz = \{\zz_1, \dots, \zz_n\}$. 
% The auxiliary model is then trained by minimizing the following loss function:
% %
% %
% %\vspace{-2mm}
% \begin{equation}
% \mathcal{L}(\theta) = -\sum_x^C c_x \log P_{\theta}(c_{x} | U_n) \nonumber
% \end{equation}
% %\vspace{-2mm}
% %
% \noindent where $P_{\theta}(c_{x} | U_n)$ is a softmax function presenting the % $P_{\theta}(t_{w_i} | w_i) = \frac{\exp (\theta_l \cdot \zz_i)}{\sum_{l'} \exp (\theta_{l'} \cdot \zz_i)} $
% %  is 
% the probability that utterance $U_n$ is assigned property $c_{x}$. 
We extract the representations from the individual layers for our layer-wise analysis (Section \ref{ssec:ul-representation}) and the entire network for the neuron-analysis (Section \ref{ssec:proxy_classifier}). 

\begin{figure}[!ht]
  \centering
  \includegraphics[width=0.9\linewidth]{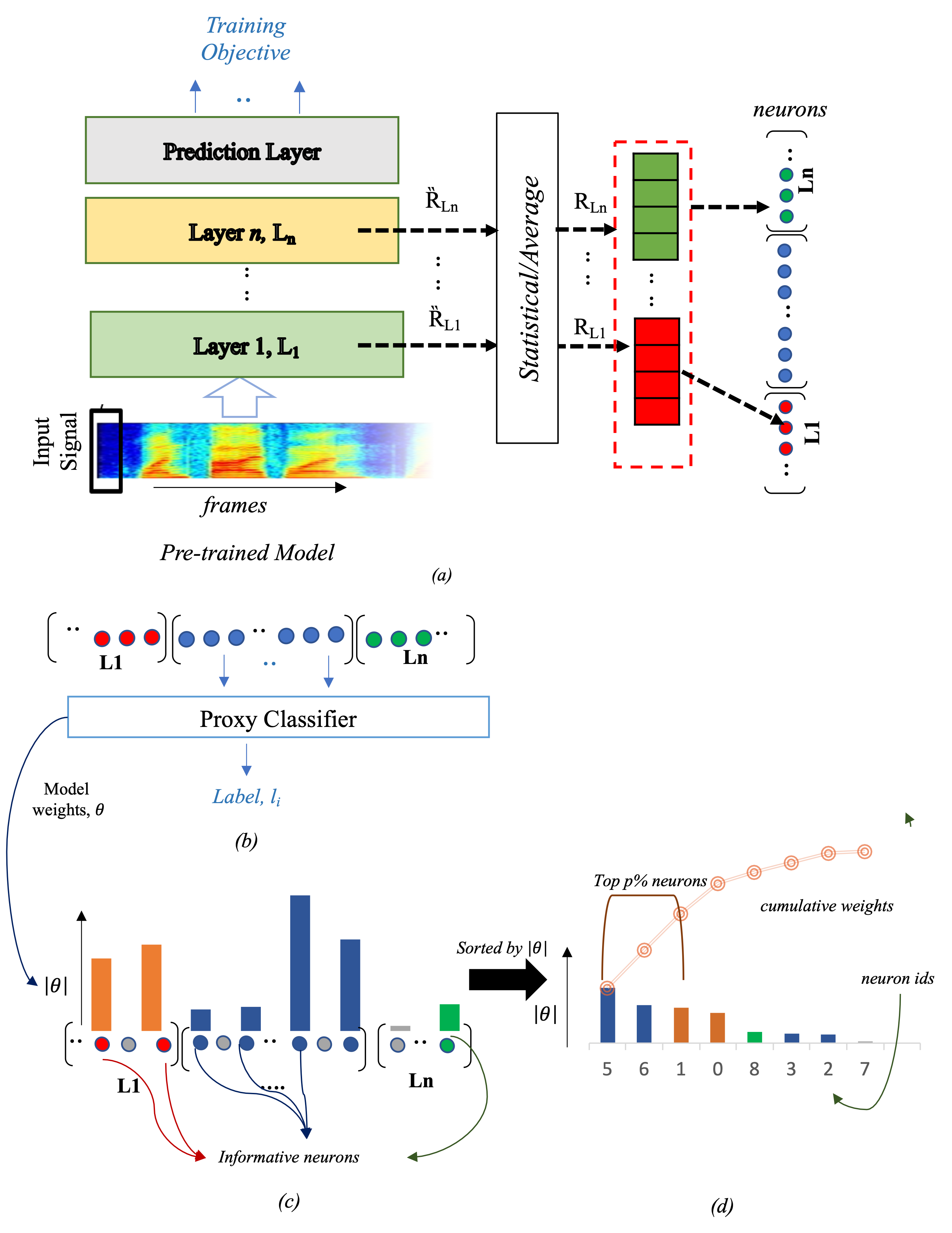}
%   \vspace{-0.3cm}
% \caption{Adding}
  \caption{Experimental pipeline of the study: In Figure \ref{fig:pipeline}(a), the process of converting frame-level representation, denoted as $\ddot{\mathcal{R}}_l$, to utterance-level representation, denoted as $\mathcal{R}_l$, using average/statistical polling is depicted. Figure \ref{fig:pipeline}(b) illustrates the proxy classifier, Figures \ref{fig:pipeline}(c) and \ref{fig:pipeline}(d) demonstrate the utilization of trained weights from the proxy classifier for ranking the neurons in the pipeline.}
  \label{fig:pipeline}
%   \vspace{-0.3cm}
\end{figure}

\paragraph{Neuron Analysis} We use the \emph{LCA} method \cite{dalvi2019one} to generate a neuron ranking with respect to the property of interest. The classifier is trained by minimizing the following loss function:
%
%\vspace{-2mm}
\begin{equation}
\mathcal{L}(\theta) = -\sum_i \log P_{\theta}(t_{u_i} | u_i)  \nonumber
\end{equation}
%\vspace{-2mm}
%
where $P_{\theta}(t_{u_i} | u_i) = \frac{\exp (\theta_l \cdot z_i)}{\sum_{l'} \exp (\theta_{l'} \cdot z_i)}$ is the probability that utterance $u_i$ is classified as property $t_{u_i}$. The weights $\theta \in \mathds{R}^{D \times |T|}$ are learned with gradient descent. Here $D$ is the dimensionality of the vector representations $z_i$ (of the utterance $u_i$) and $|T|$ is the number of properties the classifier is predicting. Given the trained weights, $\theta$ ($\theta \in \mathds{R}^{D \times |T|}$), of the classifier $\mathds{P}$, we want to extract a ranking of the $D$ neurons in the model $\mathds{M}$. For the property $t \in T$, we sort the weights $\theta_{t} \in \mathds{R}^D$ by their absolute values in descending order. Hence, the neuron with the highest corresponding absolute weight in $\theta_{t}$ appears at the top of our ranking. We consider the top $n$ neurons (for the individual property under consideration) that cumulatively contribute to some percentage of the total weight mass as \textit{salient neurons}. The complete process is illustrated in Figure \ref{fig:pipeline}.

 %  Given the trained classifier 
% %  $\theta \in \mathbb{R}^{\boldsymbol{Neu} \times C}$, 
%  the algorithm extracts a ranking of the neurons 
% % , $\boldsymbol{Neu}$, 
%  in the model based on weight distribution. 
%The entire process is presented in Figure \ref{fig:pipeline}. 

% Our  methodology serves two goals in this study: i) analyzing speech models to understand what properties are encoded in the networks, and ii) how localized or distributed these properties are across different layers in the network.
% More specifically, we probe for the following properties: i) speaker information: gender and voice identity, ii) language information: language and dialect identification, and iii) transmission channel information.

The methodology employed in this study serves two overarching objectives: first, to analyze speech models in order to gain insights into the specific properties encoded within the networks, and second, to assess the degree of localization or distribution of these properties across different layers of the network. In particular, we investigate the presence of three key properties: i) speaker information, such as gender and voice identity; ii) language information, including language and dialect identification; iii) and transmission channel information.

\subsection{Utterance-level Representation}
\label{ssec:ul-representation}
For a given temporal ordered feature sequence input, with $F$ number of frames and $D$ feature dimension ($D \times F$), we first extract the latent frame ($\ddot{\mathcal{R}}_l$) or utterance-level ($\mathcal{R}_l$) speech representation from the layers ($l$) of the pretrained neural network model $\mathds{M}$. 
Since our goal is to study utterance level representation $\mathcal{R}_l$ for a layer $l$, we aggregate the frame-level representations by passing it through a statistical/average pooling function ($\varphi^{[l]}$), $\mathcal{R}_l = \varphi^{[l]}(\ddot{\mathcal{R}}_l)$. 
% For the entire network representation, we concatenate\footnote{When training the classifier, we form a large vector of N features (N: hidden dimensions × number of
% layers) by concatenating the layers.} the layers ($l$) to obtain $ALL = \mathcal{R}_{l_1} + \mathcal{R}_{l_2} + .. + \mathcal{R}_{l_L}$. 
For the entire network representation, we concatenate\footnote{When training the classifier, we form a large vector of $N$ features ($N$: hidden dimensions $\times$ number of layers) by concatenating the layers.} the layers ($l$) to obtain $ALL = \mathcal{R}{l_1} + \mathcal{R}{l_2} + \ldots + \mathcal{R}_{l_L}$, where $L$ represents the total number of layers.

\subsection{Proxy Classifier}
\label{ssec:proxy_classifier}

%We then design proxy classifiers, $\mathds{M}_{T}$ for a task, $\mathcal{T}$. 
Given a dataset of $N$ utterances $\mathbb{U}=\{u_1, u_2, ..., u_N\}$ with a corresponding set of annotation classes $C=\{c_1, c_2, ..., c_C\}$, we map each utterance $u_i$, in the data to the latent utterance-level representations using pretrained model $\mathds{M}$. 
%As a proxy classifier, we opt for 
We use a simple %linear -- 
logistic regression model ($\mathds{P}$) trained by minimising the cross entropy loss $H$. The accuracy of the trained classifier serves as a proxy to indicate that the model representations have learned the underlying property.

%The trained classifier $\mathds{M}_{T}$, is then used to evaluate the %strength 
%quality of the input representation, by measuring the %classification performance
%classifier accuracy. The probe provides insights into the %strength 
%quality of the encoded information, yet it has the potential to identify informative neurons in the network. 

\paragraph{Neuron Analysis}  We use the weights of the trained classifier to measure the importance of neurons with respect to the understudied property. Because neurons are multi-variate in nature, we additionally use elastic-net regularization \cite{zou2005regularization}:

%To carryout the fine-grained neuron analysis, we modified the proxy classifier adding elastic-net regularization \cite{zou2005regularization}, as shown below: 

\begin{equation}
\label{eq:loss}
    \mathcal{L}(\theta)= H_{\theta}+ \lambda_1\left \|\theta \right \| _{1} + \lambda_2\left \|\theta \right \| _{2}^{2}
\end{equation}

\noindent where $\theta$ represents the learned weights in the classifier and $\lambda_1\left \|\theta \right \| _{1}$ and $\lambda_2\left \|\theta \right \| _{2}^{2}$ correspond to $L_{1}$ and $L_{2}$ regularization respectively. The combination of $L_{1}$ and $L_{2}$ regularization creates a balance between selecting very focused localised features ($L_{1}$) \emph{vs} distributed neurons ($L_{2}$) shared among many properties. 
%Using the modified loss with the $\lambda_*$ parameters, we trained the proxy classifiers to access the learned weights for measuring the importance of each neuron. 

\begin{algorithm}[h]
\label{algo:top}
\caption{Top Neuron Ranking}
\SetAlgoLined
\SetKwFunction{FMain}{TopNeurons}
    \SetKwProg{Fn}{Function}{:}{}
    \Fn{\FMain{$\theta, p$}}{
        $\theta_{top, c} \longleftarrow [ ][ ]$ \Comment{To store top neurons per class-label}
        
        \ForEach{$ class, c \in C $}
        { 
        $ tm \longleftarrow \sum_{n=1}^{N}|\theta_{n}^{c}|$ \Comment{total mass}
        
        $\theta_{s} \longleftarrow sort(\theta^{c})$ \Comment{sorted list by weight}
        
        $\theta_{cm} \longleftarrow cumulativeSum(\theta_s)$ 

        $\theta_{top, c} \longleftarrow \theta_{cm} <  p* tm$  \Comment{top neurons per class with threshold p}
        }
        $\theta_{top} \longleftarrow  \bigcup_{c=1}^{C}\theta_{top, c}$ \Comment{top neurons for all the classes}
        
        \textbf{return} $\theta_{top} $ %\; 
}
\end{algorithm}

\subsection{Selecting Salient Neurons}
\label{ssec:fg-neuron}

Our goal is to determine the relative importance of intermediate neurons in relation to specific task properties.  To achieve this, we rank the neurons based on their significance for each class label $c$. Using the trained proxy classifier $\mathds{P}$, we first arrange the absolute weight values of the neurons $|\theta_{n}^{c}|$ in descending order, and then calculate the cumulative weight vector (as illustrated in Figure \ref{fig:pipeline}c-d). The total mass is defined by the sum of the weight vector, and each element $n$ indicates the portion of the weight mass covered by the first $n$ neurons. Subsequently, we select the top p\% of neurons, corresponding to the cumulative weights that constitute a certain percentage of the total mass.
% To obtain this, we rank the neurons based on 
% their importance per class label $c$. Given the trained proxy classifier $\mathds{P}$, we first sort the absolute weight values of the neurons $|\theta_{n}^{c}|$ from $\mathds{P}$ in descending order and calculate the cumulative weight vector (as shown in Figure \ref{fig:pipeline}c-d). The sum of the weight vector defines the total mass and each element $n$ defines how much of the weight mass the first $n$ neurons cover. We then select the top p\% neurons, corresponding to the percentage of cumulative weights of the total mass.
To find the salient neurons for each class, we initialize with a small percentage ($p=0.1\%$) of the total mass. Then, the percentage is iteratively increased, while adding the newly discovered important neurons ($salient\_neurons$ $\leftarrow$ TopNeurons($\theta$,p) $\setminus$ $salient\_neurons $) to the list ordered by importance towards the task \cite{dalvi2019neurox}. The algorithm terminates when the salient neurons achieve accuracy close to the {\emph Oracle} (i.e., accuracy using the entire network) within a predefined threshold. Further details can be found in Section \ref{sec:minimalNeurons}. Finally, we combine all the selected class-wise neurons to determine the overall top neurons of the network (see Algorithm 1 for specifics).
% Finally, we combine all the selected class-wise neurons to get overall top neurons of the network (see details in Algorithm 1).
% \ref{algo:top}).

% To obtain task-wise salient neurons, an initial small percentage ($p=0.1\%$) of the total mass is used to find the salient neurons for all the classes. Then, the percentage is iteratively increased, while adding the newly discovered important neurons ($salient\_neurons$ $\leftarrow$ TopNeurons($\theta$,p) $\setminus$ $salient\_neurons $) to the list ordered by importance towards the task \cite{dalvi2019neurox}. The algorithm terminates when the salient neurons obtain accuracy close to the {\emph Oracle} (accuracy using the entire network) within a certain threshold. See \ref{sec:minimalNeurons} for details.

% \begin{algorithm}[h]
% \label{algo:top}
% \caption{Top Neuron Ranking}
% \SetAlgoLined
% \SetKwFunction{FMain}{TopNeurons}
%     \SetKwProg{Fn}{Function}{:}{}
%     \Fn{\FMain{$\theta, p$}}{
%         $\theta_{top, c} \longleftarrow [ ][ ] $ \Comment{To store top neurons per class-label}
%         % $\theta_{top, c} \longleftarrow [ ][ ]$\; \Comment{To store top neurons per class-label}
% }
% \end{algorithm}

\subsubsection{Efficacy of the Ranking}

%To 
% We evaluate the effectiveness of neuron ranking per encoded information using ablation. More specifically we select the top/bottom/random $20\%$ of neurons, while masking-out\footnote{We assigned zero to the activation of the masked neurons.} the remaining from the ranked list as prescribed in \cite{dalvi2019one}. We then re-evaluated the test set, using the  previously trained proxy classifier. 

We evaluate the effectiveness of neuron ranking for encoded information using ablation. More specifically, we select the top/bottom/random 20\% of neurons, while masking-out,\footnote{We assigned zero to the activation of the masked neurons.} the remaining from the ranked list as prescribed in \cite{dalvi2019one}. Next, we re-evaluated the test set using the previously trained proxy classifier.

\subsubsection{Minimal Neuron Set}
\label{sec:minimalNeurons}

%The neural networks are innately distributive with a handful of neurons capturing most of the information for a particular property. Our search for such specialized neuron sets lead us to believe that a minimal set of neurons -- that can regain the performance of the whole network. This %could 
%emphasises the redundant nature of the network and the importance of neuron-level analysis. 

To obtain a minimal neuron set for the proxy task, we iteratively gather an array of top $N$\% of neurons and re-train the proxy classifier with the selected neurons only. We repeat this method\footnote{Using the top 1, 5, 10, 15, 20, 25, 50 and 75 \%  neurons.} until we converge to \textit{Oracle} performance (Accuracy) i.e. with the model trained with all the neurons (`ALL') of the network. We use a threshold of $\delta=1.0$ as our convergence criteria i.e. the number of neurons sufficiently capture a property if retraining classifier, using them, results in only $1\%$ loss in accuracy. We select the neuron set with the highest accuracy close to $[Acc(ALL)\pm\delta]$. 

\subsection{Control tasks}
\label{ssec:control}

Using probing classifiers for analysis presents a potential pitfall: determining whether the accuracy of the probe reflects the properties learned in the representation or merely indicates the classifier's capacity to memorize the task \cite{hewitt2019designing, pimentel2020informationtheoretic, voita2020informationtheoretic}. To ensure that the probe's performance accurately reflects the quality of the representation in terms of encoded information, we conducted two control tasks. First, we evaluated the performance of the extracted embedding by training the classifier with randomly initialized features. Second, we assessed the probe's ability to memorize random class label assignments using a selectivity criterion \cite{hewitt2019designing}.

% A pitfall to using the probing classifiers for analysis is whether the accuracy of the probe is a true reflection of property learned in the representation or indicative of the classifier's capacity to memorize the task \cite{hewitt2019designing, pimentel2020informationtheoretic, voita2020informationtheoretic}. To ensure that probe's performance is reflective of the representation quality with respect to the encoded information, we designed the following two control tasks:  firstly, we test the performance of the extracted embedding by training the classifier with randomly initialized features, secondly, we test the ability of the probe to memorize random assignment of the class labels using selectivity criterion \cite{hewitt2019designing}. 
% The Selectivity is defined as We then calculate the performance difference of the probe, comparing the proxy task with the control task result.

The control task for our probing classifiers is defined by mapping each utterance type $u_i$ to a randomly sampled behavior $C(u_i)$ from a set of numbers ${1 . . . T}$, where $T$ is the size of the tag set to be predicted in the property of interest. The assignment is done in such a way that the original class distribution is maintained in the train-set.
%Given all the classes ($C_T$) of the proxy task, we randomly assign the training instances to $c_{i}$ ($c_{i} \in C_T$), maintaining the original class distribution in the train set. 
We compute the probing accuracy of the task and the control task. The selectivity is computed by subtracting the performance of the probe with control labels from the reported proxy task performance, $Sel_a = Acc(ALL) - Acc_{R}(ALL)$, where $R$ indicates the dataset with control labels. For more details, please refer to \cite{hewitt2019designing}.

\subsection{Redundancy}
\label{ssec:redundancy}

We conduct redundancy analysis on the results based on the concept of redundancy as described in \cite{dalvi2020analyzing}. To determine redundancy, we utilize the performance of a classifier as an indicator of task-specific knowledge learned by a layer or group of neurons. If two layers of the model achieve similar performance close to the oracle (within a certain threshold), we consider them redundant with respect to a downstream task. The oracle refers to the skyline performance of the model when the entire model is used for task prediction.

Similarly, we identify subsets of neurons as redundant with respect to a downstream task if they serve the same purpose in terms of feature-based transfer learning, meaning they can be used to train a classifier for a downstream task with similar performance close to the oracle (within a certain threshold). This redundancy arises due to over-parameterization and other training choices that encourage different parts of the models to learn similar information.

\begin{figure}[]
  \centering
  \includegraphics[width=\linewidth]{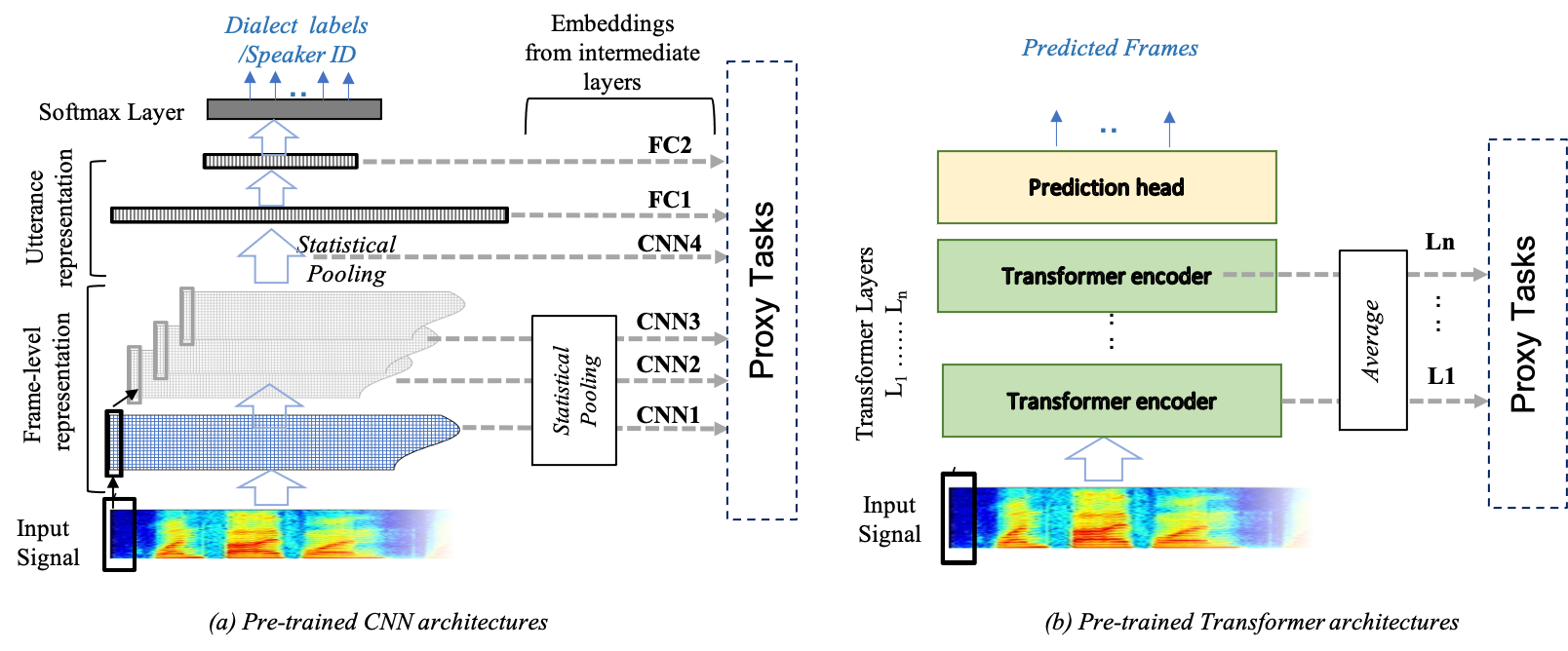}
%   \vspace{-0.3cm}
% \caption{Adding}
  \caption{Architecture of the pretrained models. Figure \ref{fig:archi}a presents the CNN and Figure \ref{fig:archi}b presents the transformer architecture. FC - fully-connected layer, CNN - convolution layer. }
  \label{fig:archi}
%   \vspace{-0.3cm}
\end{figure}

\section{Experimental Setup}
\label{sec:expsetup}

\subsection{Pretrained Models}
\label{ssec:models}

We experimented with a temporal Convolutional Neural Network (CNN) trained with two different objective functions and two Transformer architectures (refer to Sections \ref{sssec:cnn} and \ref{sssec:transformer}). The choice of architectures is motivated by the effectiveness of CNNs in modelling different speech phenomena, as well as the increasing popularity and state-of-the-art performance of Transformers in speech and language modeling.

% We experimented with temporal Convolution Neural Network trained with two different objective functions (see Section \ref{sssec:cnn}) and two %speech 
% Transformer architectures (see Section \ref{sssec:transformer}). The choice of architectures is motivated by the following: i) CNNs, due to their effectiveness in modelling different speech phenomena, and ii) Transformers, due to their increase in popularity and state-of-the-art performance in speech and language modeling.
% \nd{It would be good to say here what motivated this choice of models. Like are they state-of-the-art? and that it would be interesting to compare these two different architectures, if possible}

% For the CNN models,\footnote{See Appendix \ref{appen:cnn} for detailed model parameters.} 
% we choose two pretrained models optimized for the task of predicting: (i) Arabic dialects (resultant model refereed as ADI) and (ii) speakers (refereed as SRE).
% For the speech transformers, we opt for Mockingjay\footnote{See Appendix \ref{appen:transformer} for detailed parameters.} \cite{liu2020mockingjay} models. 

% -- a variant to language identification task, ADI, and (ii) a speaker recognition, SR task. 

\subsubsection{CNN} 
\label{sssec:cnn}

The CNN models, as illustrated in Figure \ref{fig:archi}a, comprise four temporal CNN layers followed by two feed-forward layers\footnote{Refer to \ref{appen:cnn} for detailed model parameters.}. These models are specifically optimized for two tasks: i) Arabic dialect identification (\textit{ADI}) and ii) speaker recognition (\textit{SR}). Dialectal Arabic is used in 22 countries, with over 20 mutually incomprehensible dialects and a shared phonetic and morphological inventory, making Arabic Dialect Identification particularly challenging when compared to other dialect identification tasks \cite{ali2021connecting}. The model is trained using the Arabic Dialect Identification 17 (ADI17) dataset \cite{adi17,mgb5}.
% \sout{As for such a discriminating task, Arabic is an appropriate language choice -- shared among 22 countries, with more than 20 mutually incomprehensible dialects and a common phonetic and morphological inventory.}
% \sout{Whereas the SR model (referred as \textit{SRE}),\footnote{SR is speaker recognition model and E (English) represent the language of the model training data.}}
We conducted experiments on speaker recognition task using English (SRE). We adapted the approach from \cite{shon2018frame} and trained the model using the Voxceleb1 development set, which includes data from 1211 speakers and approximately 147K utterances.

\subsubsection{Transformer}
\label{sssec:transformer}

% \sout{For the speech transformers, we opt for Mockingjay\footnote{See \ref{appen:transformer} for detailed parameters.} \cite{liu2020mockingjay} models.}

We included two transformer-encoder architectures\footnote{See \ref{appen:transformer} for detailed model parameters.} using Mockingjay \cite{liu2020mockingjay}, of which we tried two variations differing in the number of encoder layers (3 or 12 -- see Figure \ref{fig:archi}b). We refer to the former as base ($ST_{base}$) and the latter as large ($ST_{large}$) models.
% \sout{(see Figure \ref{fig:archi}b) differed by the number of transformer-encoder layers present in the model -- 3 and 12 encoder layer architecture, referred to as base ($ST_{base}$) and large ($ST_{large}$) models.} \sout{The models are trained to predict \sout{the current} \hll{a} frame while jointly conditioning on \sout{both} \hll{the} past and future contexts.}
The base model ($ST_{base}$) is trained using Mel features as the reconstruction target, whereas for the large model ($ST_{large}$) we use a linear-scale spectrogram as the reconstruction target.
%the reconstruction target was a linear- scale spectrogram. 
The models were trained using the LibriSpeech corpus:train-clean-360.\footnote{\url{https://www.openslr.org/resources/12/train-clean-360.tar.gz}}

\subsection{Proxy Tasks}
\label{ssec:probing}

We conducted experiments in our study using the following proxy tasks: i) Gender Classification, ii) Speaker Verification, iii) Language Identification, iv) Regional Dialect Classification, and v) Channel Classification. Below, we provide brief details related to each of these tasks.

\subsubsection*{T1 - Gender Classification (GC)}

\begin{figure} [!htb]
\centering
\scalebox{0.7}{
\includegraphics[width=\linewidth]{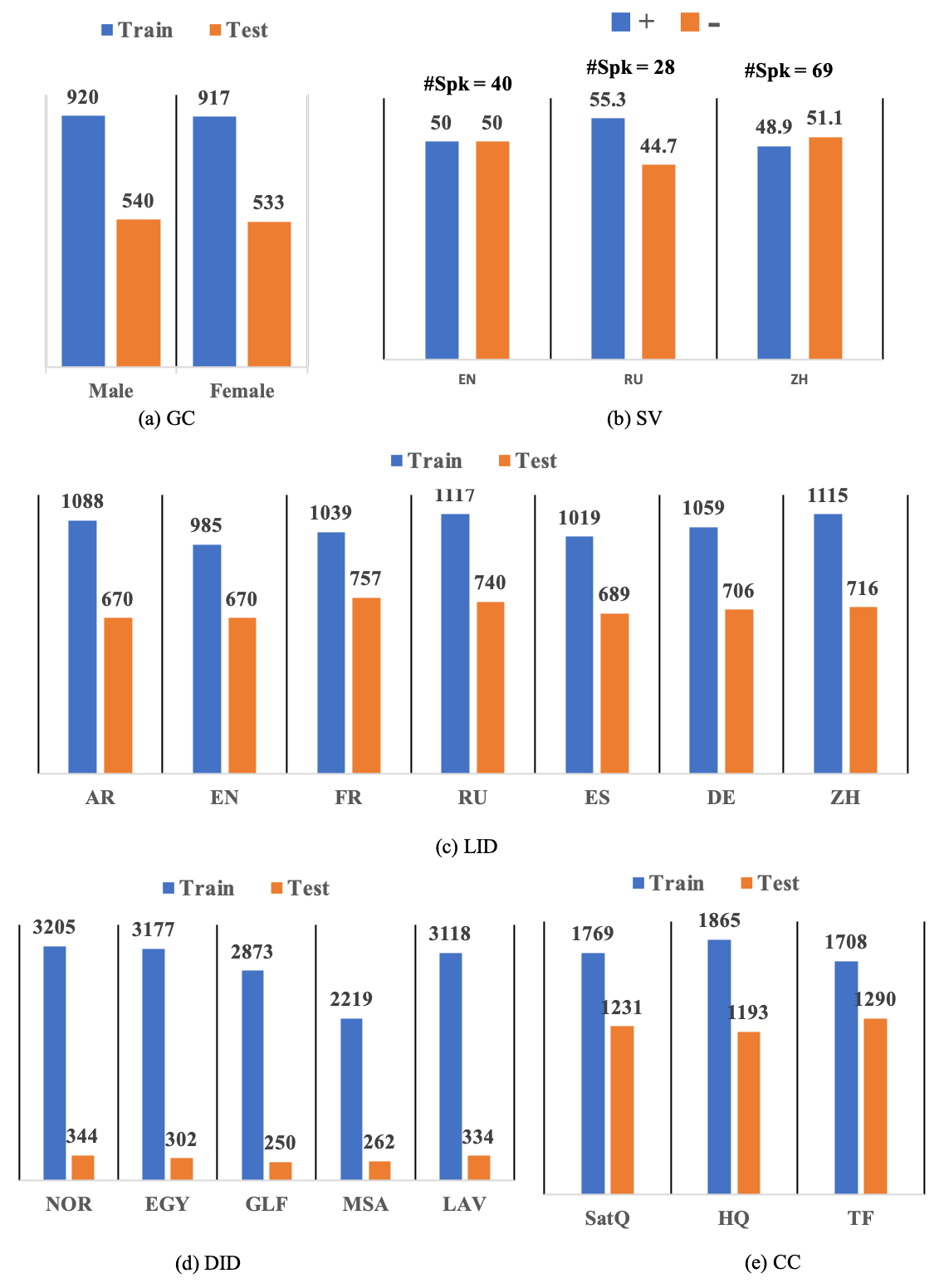}
}
\caption{Train and Test Data distribution for proxy classification tasks. 
 Figure \ref{fig:data_dist}(a) shows data distribution for GC -- Gender Classification (Task1). Figure \ref{fig:data_dist}(b) SV -- Speaker Verification (Task2): the blue bar represents the positive (from the same speaker) label and the orange is the negative label (from different speaker pairs). $\#Spk$ represents the total number of speakers in each class label ($+/-$) for English EN, Russian RU, and Chinese ZH data. Figure \ref{fig:data_dist}(c) LID -- Language Identification (Task3); Figure \ref{fig:data_dist}(d): DID -- Regional Dialect Identification (Task4); and Figure \ref{fig:data_dist}(e): CC -- Channel Classification (Task5).
 For Figure \ref{fig:data_dist}(a, c-e), the x-axis indicates the class labels whereas the y-axis of each bar chart represents the frequency of the corresponding class. 
 }

\label{fig:data_dist}
% \vspace{-0.5cm}
\end{figure}

For the gender classification task, we trained proxy classifiers using the VoxCeleb1-test \cite{nagrani2017voxceleb} (English) dataset, which includes videos of celebrities from different ethnicities, accents, professions, and ages.
 We used gender-balanced train and test sets with no overlapping speakers. The detailed label distribution for the task is shown in Figure \ref{fig:data_dist}(a).

\subsubsection*{T2 - Speaker Verification (SV)}

We conducted two experiments for the speaker verification task, focusing on two aspects. Firstly, we investigated the presence of voice identity information in different layers of the network. Secondly, we aimed to identify the minimum subset of neurons that could effectively encode voice identity. For the first experiment, we performed a ``generic speaker verification'' task using pairs of input signals, where we verified if two utterances are from the same or different speaker. For each input signal pair, we extracted length-normalized embeddings from individual layers of the network, as well as their combination (referred to as ``ALL''). Subsequently, we computed the cosine similarity between the pairs to analyze the results.

% For the speaker verification task, we conducted two experiments, studying (i) whether the information is present in the layers of the network and (ii) finding a minimal neuron subset that can encode voice identity.

% For the first experiment, we performed a `generic speaker verification' task on pairs of input signals. Given two utterances, the task is to verify if they are from the same or different speaker. 
% To do this, we 
% For each input signal pair, we first extracted length normalized embeddings from individual layers and their combination (`ALL')
% % , of the pretrained models 
% and then computed the cosine similarity between pairs. 

% The second experiment, for the neuron-level analysis, we trained a proxy classifier -- for the speaker recognition task -- using embeddings from the layers of the pretrained models. We selected the minimal neurons, using the algorithm described in Section \ref{sec:minimalNeurons} and then used it to test our verification pairs.
% For training the proxy speaker recognition model, we used \textit{VoxCleleb2}-test set \cite{chung2018voxceleb2} containing $118$ speakers, with $4,911$ videos and $36,237$ utterances.  

In the second experiment, we conducted a neuron-level analysis by training a proxy classifier for speaker recognition using embeddings from layers of pretrained models. We employed an algorithm detailed in Section \ref{sec:minimalNeurons} to select the minimal neurons, which were then used to test our verification pairs. The proxy speaker recognition model was trained on the \textit{VoxCeleb2}-test set \cite{chung2018voxceleb2}, which consists of $118$ speakers, $4,911$ videos, and $36,237$ utterances.

In our speaker verification tasks, we utilized a multi-lingual subset of the \textit{Common Voice} dataset \cite{ardila2019common}, as well as the official verification test set (Voxceleb1-tst) from \textit{VoxCeleb1}\footnote{In this case, we used the official verification pairs to evaluate.} \cite{nagrani2017voxceleb}.
The \textit{Common Voice corpus} is an extensive collection of over $2,500$ hours of speech data in approximately $39$ languages.\footnote{last accessed: April 10, 2020.} It was collected and validated through a crowdsourcing approach, making it one of the largest multilingual datasets available for speech research. The data was recorded using a website or an iPhone application provided by the Common Voice project. For our experiments, we selected three languages: English (EN), Russian (RU), and Chinese (ZH), and used the Voxceleb1-tst dataset along with a randomly selected subset of Common Voice data, totaling approximately $4$ hours.

 % The \textit{Common Voice corpus} contains more than $2,500$ hours of speech data from $\approx39$ languages,\footnote{last accessed: April 10, 2020.} collected and validated via a crowdsourcing approach. This is one of the largest multilingual datasets available for speech research recorded using a website or an iPhone application available from the Common Voice project. We experimented using three languages including English (EN), Russian (RU) and Chinese (ZH) using the Voxceleb1-tst and a subset ($\approx$4 hours) of randomly selected data from Common Voice.
%datasets.  We constructed the RU and ZH verification pair trials by randomly picking up utterance pairs from speakers with the same gender, maintaining a balanced distribution between positive and negative targets. Details of the verification pairs are given in Figure \ref{fig:data_dist}(b).

\subsubsection*{T3 - Language Identification (LID)} 

% For the language identification task, we designed classifiers for discriminating between the $7$ languages selected from the Common Voice dataset. The language subset used for this study includes: Arabic (AR), English (EN), Spanish (ES), German (DE), French (FR), Russian (RU) and Chinese (ZH). The distribution of the datasets for training and testing the classifiers are shown in Figure \ref{fig:data_dist}(c). 

We developed classifiers to discern between 7 different languages from the Common Voice dataset for the language identification task. The languages included in our study are Arabic (AR), English (EN), Spanish (ES), German (DE), French (FR), Russian (RU), and Chinese (ZH). The distribution of the datasets for training and testing the classifiers can be seen in Figure \ref{fig:data_dist}(c).

\subsubsection*{T4 - Regional Dialect Classification (DID)}

% For training the regional dialect classification model, we used the Arabic ADI-5 dataset \cite{ali2017speech}, which is composed of the following five dialects:
% Egyptian (EGY), Levantine (LAV), Gulf (GLF), North African Region (NOR) and Modern Standard Arabic (MSA).  The dataset contains satellite cable recording (SatQ) in the official training split and high-quality (HQ) broadcasts videos for development and test set. For the classification, we used the balanced train set to design the proxy task and tested using the test split. 
% Details of the class distribution is reported in Figure \ref{fig:data_dist}(d).

To train our regional dialect classification model, we utilized the Arabic ADI-5 dataset \cite{ali2017speech}, which comprises five dialects: Egyptian (EGY), Levantine (LAV), Gulf (GLF), North African Region (NOR), and Modern Standard Arabic (MSA). The dataset includes satellite cable recordings (SatQ) in the official training split, as well as high-quality (HQ) broadcast videos for the development and test sets. We designed the proxy task using the balanced train set and evaluated the model using the test split. Figure \ref{fig:data_dist}(d) provides detailed information on the class distribution.

\subsubsection*{T5 - Channel Classification (CC)}

% We used the ADI-5 \cite{ali2017speech} dataset and CALLHOME\footnote{\url{https://catalog.ldc.upenn.edu/LDC97S45}} \cite{kumar2014translations,billa1997multilingual} dataset to probe the Channel Classification task. Our multi-class classifier with labels indicate the input signal quality as Satellite recording (SatQ), high quality archived videos (HQ)  {\em or} Telephony data (TF). The SatQ data was built using the ADI-5 train data. The HQ data was build using the ADI-5 dev and test sets. For the TF data, we upsampled the CALLHOME data to 16kHz sampling rate. We used a VAD to split the conversation into speech segments and then randomly selected segments with a duration greater than 2.5 secs.

In our study on Channel Classification, we utilized two datasets: ADI-5 \cite{ali2017speech} and CALLHOME\footnote{\url{https://catalog.ldc.upenn.edu/LDC97S45}} \cite{kumar2014translations,billa1997multilingual}. Our multi-class classifier assigns labels to indicate the input signal quality, which can be Satellite recording (SatQ), high-quality archived videos (HQ), or Telephony data (TF). To create the SatQ data, we used the ADI-5 train data, while the HQ data was built using the ADI-5 development and test sets. For the TF data, we upsampled the CALLHOME data to a 16kHz sampling rate. We employed a Voice Activity Detector (VAD) to segment the conversations into speech segments and then randomly selected segments with a duration exceeding 2.5 seconds. To conduct the proxy task, we randomly selected balanced samples from each class and divided them into train and test sets using a 60-40\% split for our experiment. The dataset distribution is visualized in Figure \ref{fig:data_dist}(e).\footnote{A similar pattern is observed when experimenting using just SatQ and HQ labels and removing the upsampled TF. For brevity we are only presenting the experiments with SatQ, HQ and TF.}

\subsection{Classifier Settings}
\label{ssec:model_setting}

We trained logistic regression models with elastic-net regularization, using the Adam optimizer with a default learning rate, for 20 epochs and a batch size of 128. The classifier was trained by minimizing the cross-entropy loss.
%For simplicity, 
We used fixed values for $\lambda_*$ ($\lambda_*=0$ and $\lambda_*=1e-5$) in our experiments.\footnote{In our preliminary results, we found no significant difference in neuron distribution between the $\lambda_=0$ and $\lambda_*=1e-5$.}
 % We presented our distribution analysis using $\lambda_*=0$. We keep tuning for the values of $\lambda_*$ for future.
% } 
% regularization weights. 

% $\lambda_*$\footnote{We did two sets of experiments, $\lambda_1, \lambda_2=1e-5$ and $\lambda_1, \lambda_2=0$ i.e. without regularization. We observed similar performance when probed for overall representation and comparable in class-wise neuron distribution. Therefore, for brevity, we are presenting results with $\lambda_*=0$.} regularization weights. 
% @Nadir: lamda should not effect result accuracy, but distribution onky.
% \nd{I would remove the regularization part from the paper. Your argument that regularization gives the same accuracy as using no regularization is flawed. The purpose of regularizer is to get a good mix of spiky and groups of neurons. }

% \subsection{Evaluation Measures}
% As an evaluation measure we report Accuracy for the proxy task. Given the simplicity of metric interpretibility and balanced class-labels for each proxy task, accuracy seems to be an effective measure. As for the speaker verification (SV), we report Equal Error Rate (EER) -- measuring the value at which the false-reject (miss) rate equals the false-accept (false-alarm) rate. 
\section{Layer-wise Analysis}
\label{sec:resultlc}

% We first discuss the results from training layer-wise proxy classifiers, addressing the following questions: i) does the end-to-end speech model capture different properties? ii) which parts of the network predominantly learn this property? We use several reference points to make observations.

In our initial analysis, we focus on the outcomes obtained from training layer-wise proxy classifiers, aiming to address two key questions: i) does the end-to-end speech model capture distinct properties, and ii) which specific components of the network primarily contribute to learning these properties? To gain insights, we utilize multiple reference points for observations.

% which parts of the network predominantly learns this property? We use several reference points to draw observations. %We compare against majority baseline and control tasks to establish that the network has acquired certain knowledge. We use oracle performance (classifier trained using all the network layers (`ALL')) to evaluate presence of a knowledge at different layers.

%We compare the results with the majority baseline (assigning the most frequent class for each input), and with the oracle (classifier trained using all the network layers (`ALL')) along with other control tasks mentioned in Section \ref{ssec:ns1}.

\subsection{Majority Baseline, Oracle and Control Tasks} 
\label{ssec:ns1}

% We compare against the majority baseline assigning most frequent class and control tasks to establish that the network has acquired certain knowledge. 

Table \ref{tab:task-baseline} presents the results of various tasks. We start by comparing against the majority baseline (`Maj-C'), which predicts the most frequent class in the training data for all instances in the test data. To assess the presence of knowledge in our network, we train a classifier on the concatenation of all network layers (`ALL').

Our observations show that the classifier trained on the feature vectors generated from the network (`ALL') outperforms the majority baseline (`Maj-C') significantly, indicating that our network has acquired meaningful knowledge regarding the understudied properties. To further confirm that our probe is not simply memorizing, we compare against classifiers trained using random vectors (`R.INIT') and compute selectivity results \cite{hewitt2019designing}. Notably, the classifier trained on random vectors performs similarly or worse than the majority baseline. The high selectivity numbers (`$Sel_a$') provide evidence that the representations generated by our model truly reflect the acquired properties and are not just artifacts of memorization. With the effectiveness of our results established, we can now conduct a detailed discussion of the individual properties and perform a comprehensive layer-wise analysis to gain a deeper understanding of our findings.

\begin{table}[!htb]
\centering

\scalebox{0.7}{

\begin{tabular}{|l|c|c|c|c|} 
\hline
\multicolumn{1}{|l|}{} & ADI & SRE & $ST_{base}$ & $ST_{large}$ \\ 
\hline\hline
$\#Neurons $& 11100 & 11100 & 2304 & 9216 \\ 
\hline\hline
\multicolumn{5}{|c|}{T1: GC | labels\# 2} \\ 
\hline\hline
 & \multicolumn{1}{c|}{ADI} & \multicolumn{1}{c|}{SRE} & \multicolumn{1}{c|}{$ST_{base}$} & \multicolumn{1}{c|}{$ST_{large}$} \\ 
\hline\hline
$Acc$ (Maj-C) &\multicolumn{4}{c|}{56.70} \\ 
\hline
$Acc$ (ALL) & 98.20 & 96.79 & 99.16 & 98.14 \\ \hline
$Acc$ (R.INIT) & 68.14 & 68.14 & 56.17 & 56.60 \\\hline
$Sel_a$ & 42.78  & 67.28 & 52.83 & 72.53  \\\hline
\end{tabular}
\quad
 \begin{tabular}{|l|c|c|c|c|} 
\hline
\multicolumn{1}{|l|}{} & ADI & SRE & $ST_{base}$ & $ST_{large}$ \\ 
\hline\hline
$\#Neurons $& 11100 & 11100 & 2304 & 9216 \\ 
\hline\hline
\multicolumn{5}{|c|}{T3: LID | labels\# 7} \\ 
\hline\hline
 & \multicolumn{1}{c|}{ADI} & \multicolumn{1}{c|}{SRE} & \multicolumn{1}{c|}{$ST_{base}$} & \multicolumn{1}{c|}{$ST_{large}$} \\ 
\hline\hline
$Acc$ (Maj-C) &\multicolumn{4}{c|}{14.96} \\ 
\hline
$Acc$ (All) & 86.00 & 76.01 & 57.35 & 76.24 \\ 
\hline
$Acc$ (R.INIT) & \multicolumn{1}{c}{13.20} & \multicolumn{1}{c|}{13.20} & \multicolumn{1}{c|}{15.58} & \multicolumn{1}{c|}{14.23} \\ 
\hline 
$Sel_a$ & \multicolumn{1}{c|}{75.69} & \multicolumn{1}{c|}{69.20 } & \multicolumn{1}{c|}{41.18 } & \multicolumn{1}{c|}{61.76} \\ \hline
\end{tabular}
}
\medskip
\scalebox{0.7}{
\begin{tabular}{|l|c|c|c|c|} 
\hline
\multicolumn{5}{|c|}{T4: DID | labels\# 5} \\ 
\hline\hline
 & \multicolumn{1}{c|}{ADI} & \multicolumn{1}{c|}{SRE} & \multicolumn{1}{c|}{$ST_{base}$} & \multicolumn{1}{c|}{$ST_{large}$} \\ 
\hline\hline
$Acc$ (Maj-C) &\multicolumn{4}{c|}{23.06} \\ 
\hline
$Acc$ (ALL) & 55.63 & 39.12 & 36.66 & 39.22 \\ 
\hline
$Acc$ (R.INIT) & \multicolumn{1}{c|}{20.24} & \multicolumn{1}{c|}{20.24} & \multicolumn{1}{c|}{16.70} & \multicolumn{1}{c|}{22.45} \\ 
\hline
$Sel_a$ & 36.7 & 16.89 & 13.34 & 19.20 \\ \hline
\end{tabular}
% }
\quad
% \hfil
% \scalebox{0.8}{

\begin{tabular}{|l|c|c|c|c|} 
\hline
 \multicolumn{5}{|c|}{T5:CC | labels\# 3} \\ 
\hline\hline
 & \multicolumn{1}{c|}{ADI} & \multicolumn{1}{c|}{SRE} & \multicolumn{1}{c|}{$ST_{base}$} & \multicolumn{1}{c|}{$ST_{large}$} \\ 
\hline\hline
$Acc$ (Maj-C) &\multicolumn{4}{c|}{32.12} \\ 
\hline
$Acc$ (ALL) & 93.93 & 85.51 & 86.80 & 96.55 \\ 
\hline
$Acc$ (R.INIT) &26.52 & 28.54 & 37.74 & 37.32   \\ 
\hline
$Sel_a$ &  63.81 & 77.65 & 68.17 & 83.76 \\ \hline

\end{tabular}
}
\caption{Reported accuracy (Acc) for the proxy Tasks T1: GC, T3:LID, T4:DID and T5:CC using majority baseline (Maj-C), Oracle (ALL: with all neurons), random initialisation (R.INIT) of neuron's weight and selectivity ($Sel_a$). | labels\# represent the number of labels in the task. Reported performance are averaged over 5 runs. The standard deviation for the ALL results are presented in Table \ref{tab:task-baseline-std}. }
\label{tab:task-baseline}
\end{table}

\subsection{Encoded Properties}
\label{t1layer}

% We now present our layer-wise results. Remember that we train individual classifiers over the feature vector generated from each layer individually. This enables us to carry out a layer-wise analysis of the network. 

In this section, we present our findings from the layer-wise analysis. We trained individual classifiers using the feature vectors obtained from each layer separately. This approach allows us to conduct a comprehensive examination of the network on a per-layer basis.

\subsubsection*{T1: Gender Classification (GC)}

%\begin{figure}[!ht]
\begin{figure}[!htb]
  \centering
  \scalebox{0.9}{
  \includegraphics[width=\linewidth]{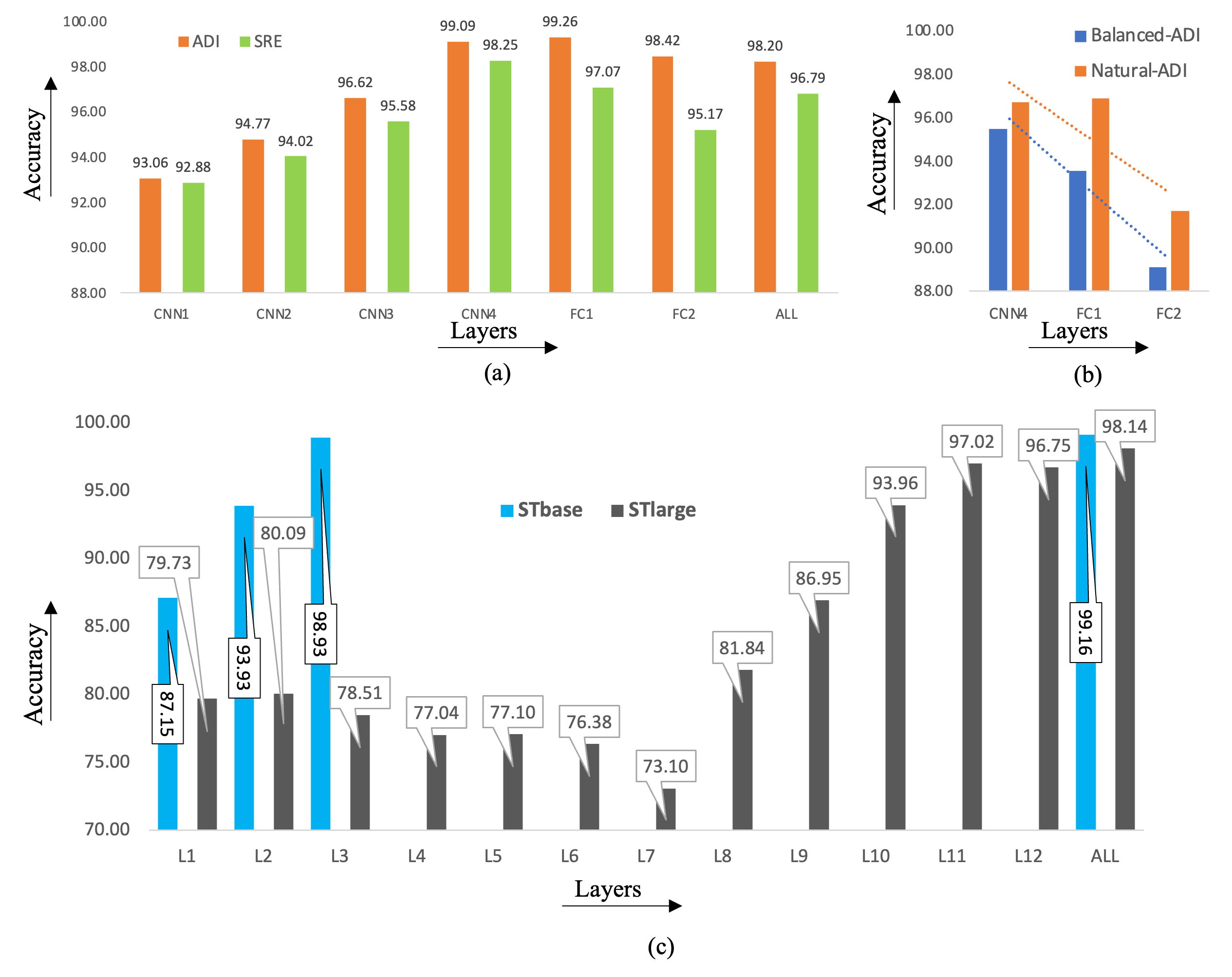} %gender_classification_layers.png
  }
%   \vspace{-0.3cm}
  \caption{Reported \textbf{accuracy}, for proxy Task Gender classification T1:GC using intermediate network layers. Figure \ref{fig:speaker_info_gc}a presents the layer-wise performance of ADI and SRE models and Figure \ref{fig:speaker_info_gc}b presents results for the last 3 layers from a similar ADI architecture, trained with balanced (\textcolor{blue}{Balanced-ADI}) and natural  (\textcolor{orange}{Natural-ADI}) distribution of gender labels for 4 dialects.
  Figure \ref{fig:speaker_info_gc}c presents the layer-wise accuracy for $ST_{b
  ase}$ and $ST_{large}$. Reported performance averaged over 5 runs. Majority baseline (assigning most frequent class): 56.70 accuracy.}
  \label{fig:speaker_info_gc}
%   \vspace{-0.3cm}
\end{figure}

Figure \ref{fig:speaker_info_gc} reveals that while the upper layers of the network acquire the most knowledge regarding gender, other layers also capture sufficient information, indicating that the property is distributed across the layers. Notably, we observed that the model trained for the ADI task appears to be more susceptible to gender information compared to models trained for other downstream tasks. We hypothesize that this could be attributed to \textit{representation bias} \cite{mehrabi2021survey}, wherein the ADI model exhibits a bias towards higher-pitched and breathier voices, which are predominantly female, because of the substantial gender imbalance and lack of variability in the training data, with fewer than one-third of speakers being female. On the other hand, the other pretrained models, such as SRE and transformers, do not demonstrate this bias.

% The results presented in Figure \ref{fig:speaker_info_gc} show that while upper layers in the network acquire most knowledge of the gender, other layers also capture the information sufficiently indicating the property 
% is distributive. Comparing the pretrained models, we noticed the model trained towards the ADI task is more susceptible to gender information than those trained towards other downstream tasks. We speculate that the ADI model shows \textit{representation bias} \cite{mehrabi2021survey} with the higher-pitched and breathier (mainly female) voices, due to a high gender imbalance and insufficient variability in the training data for the ADI model (with $< \frac{1}{3}$ female speakers). 
% In comparison, the other studied pretrained (SRE and transformers) models do not exhibit this problem.

To further investigate this hypothesis, we trained two dialect identification models,\footnote{Using the same architecture but with a subset of data from 4 dialects as class labels} while maintaining (i) the natural distribution of gender data among dialects,\footnote{Following the percentage of male-female distribution in the original data classes} and (ii) a balanced distribution. We then probed these models for gender information using the same approach. Our findings (Figure \ref{fig:speaker_info_gc}b) reveal that the upper (task-oriented) layers are less susceptible to gender information with the balanced-ADI compared to the natural-ADI. This illustrates how data imbalance in gender representation can impact the performance of speech models, as demonstrated in previous studies such as Speaker Recognition models in \cite{6613025}. Our layer- and neuron-wise analyses (Section \ref{ssec:ns2}) identify specific components of the network that exhibit such bias, reflecting the usefulness of our study for addressing bias in the network, such as imbalanced gender representation or other properties. Further exploration in this area is left for future research.

\textbf{Redundancy:} Next, we will illustrate the distribution of redundant information in the network. As a reminder, task-specific redundancy, as highlighted in \cite{dalvi2020analyzing}, refers to the parts or features of the network that are redundant with respect to a downstream task. Our findings reveal that, for the CNN architecture, the layers above CNN4 exhibit redundancy in relation to the gender classification task. In other words, the higher layers do not possess features that significantly improve performance on this task beyond 1\% compared to the best performing layer. This observation holds true for both SRE and ADI models.

\subsubsection*{T2: Speaker Verification (SV)}
\label{t2layer}

\begin{figure}[!ht]
  \centering
  \scalebox{0.8}{
  \includegraphics[width=\linewidth]{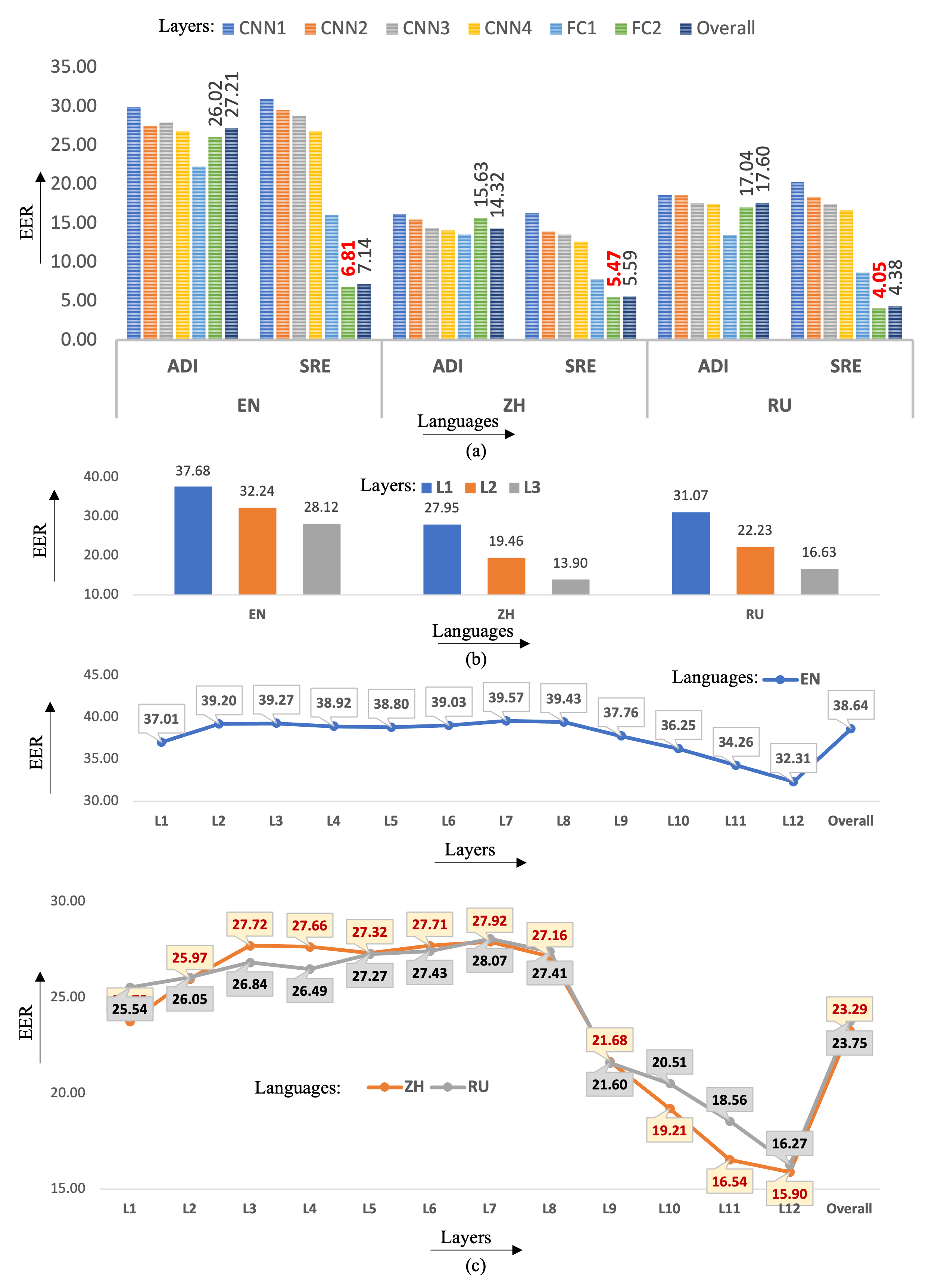} %sv_layers.png
  }
%   \vspace{-0.3cm}
  \caption{Reported \textbf{Equal Error Rate} (EER) for proxy Speaker verification, Task T2:SV using intermediate network layers. Figure \ref{fig:speaker_info_sv}a presents the layer-wise performance of ADI and SRE models for EN, ZH and RU languages, the best EER are represented in \textcolor{red}{red} color. Figure \ref{fig:speaker_info_sv}b-c presents the layer-wise EER for $ST_{base}$ and $ST_{large}$. For a better visibility, Figure \ref{fig:speaker_info_sv}c is divided into two graphs for the reported languages (EN - blue, 
ZH - orange and RU - grey curve).
  EER value, the lower the better.}
  % Reported for English - EN, Chinese - ZH and Russian - RU. 
  % }
  \label{fig:speaker_info_sv}
%   \vspace{-0.3cm}
\end{figure}

We observed that the majority of the models learn speaker-invariant representations (T2:SV -- see Figure \ref{fig:speaker_info_sv}), resulting in poor performance in the speaker verification task. This demonstrates the model's robustness towards unknown speakers and its generalization ability. When comparing the performance of the networks, we noticed that only the speaker recognition model (SRE) learned voice identity in the final layers of the network. We also observed a similar pattern when testing the SRE model on other languages, such as Chinese and Russian, indicating the domain and language-agnostic capability of the SRE model (See Figure\ref{fig:speaker_info_sv}).

\subsubsection*{T3: Language Identification (LID)}
\label{t3layer}

% We found that language markers are encoded in the upper layers (predominantly in the second last layer) in most of the studied models (Figure \ref{fig:language_info_lid}a). The upper layers have been seen to contain only task-related information (as seen in T2:SV and later in T4:DID). Hence, the presence of language information in these layers indicates their importance for discriminating between speaker or dialects. Additionally, our result emphasises that the self-supervised models also capture language information as subordinate information when encoding the input signal.

\begin{figure}[!htb]
  \centering
  \scalebox{1.0}{
  \includegraphics[width=\linewidth]{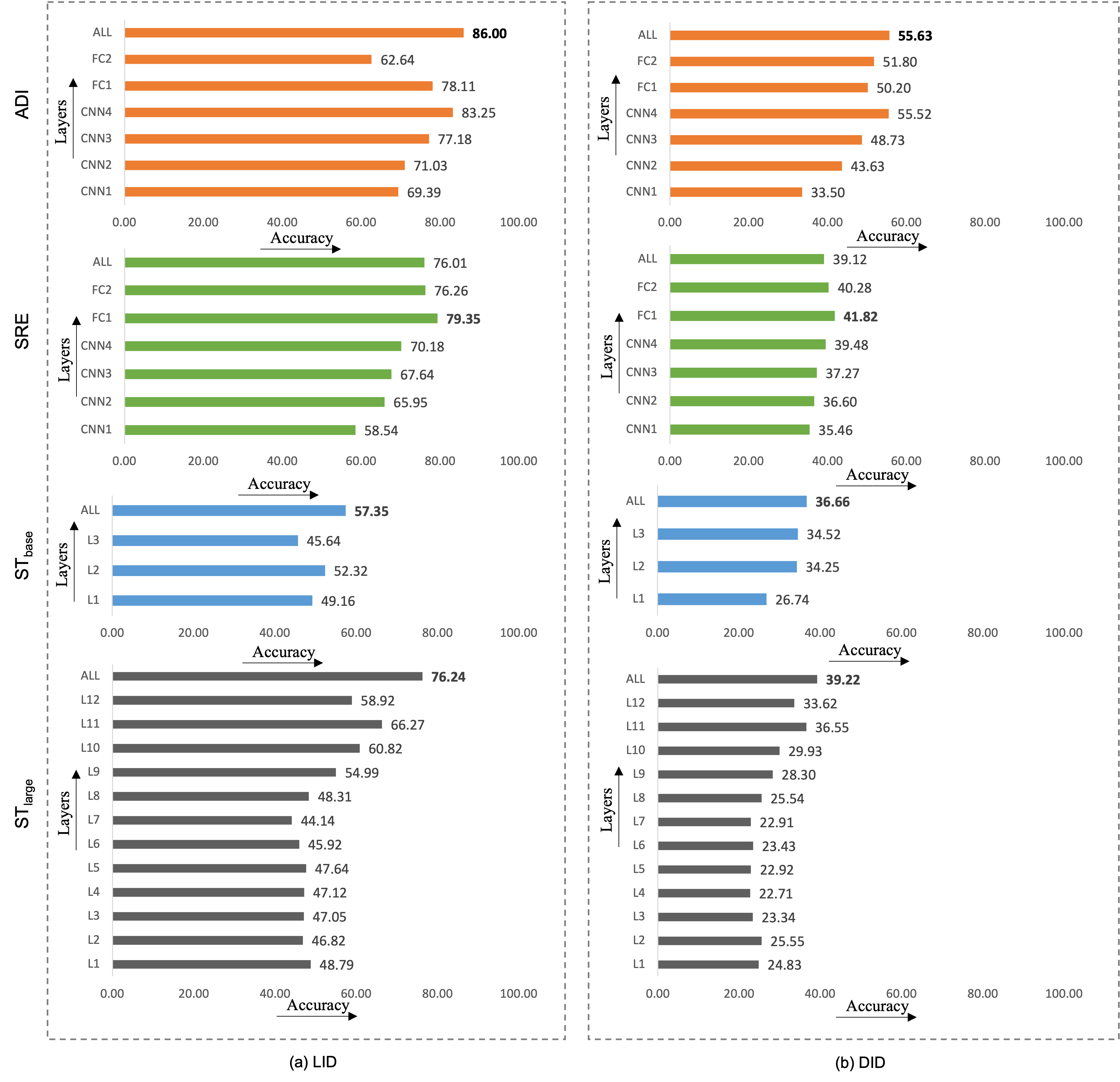}
  }
%   \vspace{-0.3cm}
  \caption{Reported \textbf{Accuracy }for proxy Tasks T3:LID and T4:DID using intermediate network layers. Figure \ref{fig:language_info_lid}a presents the layer-wise performance of all the pretrained models for the language identification task and Figure \ref{fig:language_info_lid}b presents the layer-wise accuracy for the dialect identification task. Reported performance averaged over 5 runs. Majority baseline: T3 - 14.96 and T4 - 23.06 accuracy.}
  \label{fig:language_info_lid}
%   \vspace{-0.3cm}
\end{figure}

Our findings reveal that language markers are predominantly encoded in the upper layers of most of the studied models, particularly in the second-to-last layer (as shown in Figure \ref{fig:language_info_lid}a). The upper layers primarily contain task-related information, as evidenced in T2:SV and later in T4:DID. Therefore, the presence of language information in these layers suggests their significance in discriminating between speakers or dialects. Furthermore, our results highlight that self-supervised models also capture language information as subordinate information during the encoding of the input signal.

When comparing architectures, we noticed that the CNN outperformed the transformer in capturing language representations (discussed further in detail in Section \ref{lang_neuron}). Additionally, we observed that the ADI model demonstrated significantly better performance in our probing task, particularly when using the CNN4 layer. This highlights that the model, which is originally designed to distinguish between dialects, is also proficient in discriminating languages and can be effectively fine-tuned for language identification tasks.

\subsubsection*{T4: Regional Dialect Identification (DID)}
\label{t4layer}

During our investigation of regional dialectal information, we discovered that most of the networks failed to capture the discriminating properties of dialectal information, as illustrated in Figure \ref{fig:language_info_lid}b. This reflects the complexity of distinguishing between different dialects. However, we observed that the task-specific model ADI was able to successfully capture the dialectal variation in the upper layers (CNN4-FC2) of the network. This suggests that the original pretrained models do not capture sufficient information for this complex task and that task-specific supervision is necessary to achieve accurate results, which are then preserved in the upper layers of the model.

\subsubsection*{T5: Channel Classification (CC)}
\label{t5layer}

% Similar to the gender information task (T1:GC), we observed that channel information (see Figure \ref{fig:ch_info_lid}) is omnipresent in the network. As we can see that all layers of the network perform consistently high on this task. These results are indicative of the network's ability to generalize on the mixed data with varying environments (e.g. variation in: microphones, source of data - broadcast TV, YouTube, among others).   

\begin{figure}[!ht]
  \centering
  \scalebox{0.8}{
  \includegraphics[width=\linewidth]{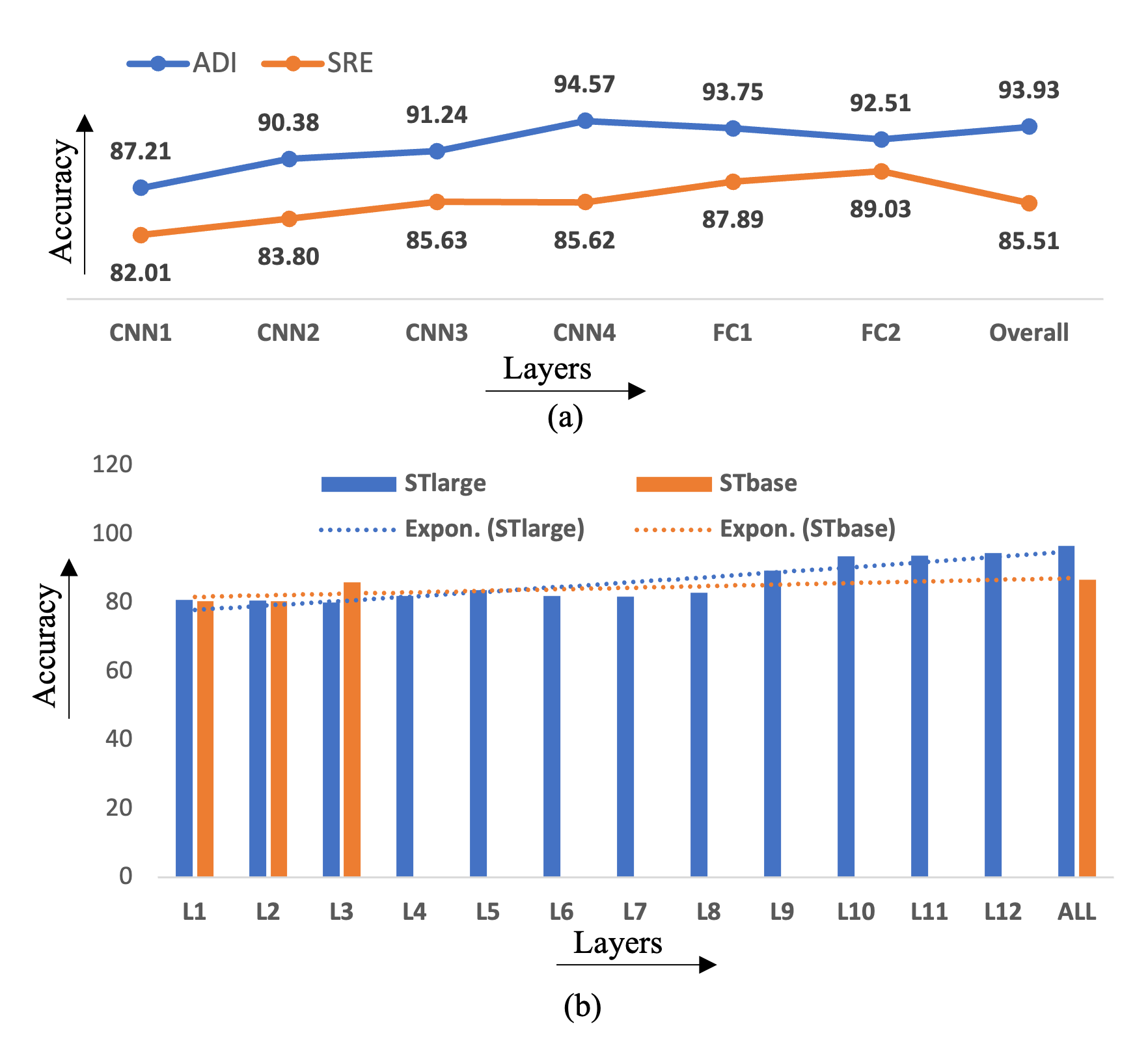}
  }
  \vspace{-0.3cm}
  \caption{Channel classification layer-wise \textbf{Accuracy}. Reported performance averaged over 5 runs. The dotted lines (Figure \ref{fig:ch_info_lid}b) show the trend-line of the graphs. Majority baseline (assigning most frequent class): 32.12 accuracy.}
  \label{fig:ch_info_lid}
  \vspace{-0.5cm}
\end{figure}

Similar to the gender information task (T1:GC), we observed that channel information (as shown in Figure \ref{fig:ch_info_lid}) is consistently present throughout all layers of the network. This indicates that the network is capable of generalizing well on mixed data that includes varying environments, such as differences in microphones and data sources (e.g. broadcast TV, YouTube, among others). The network's ability to perform consistently across all layers on this task demonstrates its robustness in handling data with diverse channel characteristics.

% -- %However, these results are not clearly indicating whether the model has learned the desired property or is just discriminating the environmental factors.
%This capability of model to capture channel information, can be potentially misleading as to whether the model has learned the desired property or just discriminating the environmental factors.
% This type of misinterpretation occurs mainly when incorporating data with varying environments (e.g. variation in: microphones, source of data - broadcast TV, YouTube, among others).  

\subsection{Summary}
\label{kolayer}

% Our layer-wise analysis shows that: i)  channel and gender information are encoded and distributed throughout the pretrained networks, ii) the gender information is redundantly encoded in the ADI and SRE pretrained models, iii) ADI network is more prone to gender-based representation bias compared to others, iv) all the pretrained model (apart of SRE) learns speaker-invariant representation, v) the voice identity is encoded only in the upper layer of SRE model, vi) the language property is captured in the upper layers of the pretrained models, 
% vii) pretrained models trained towards the task of dialect identification are a better choice to transfer knowledge for language identification downstream task, and viii) unlike the language property, the regional dialectal information are only captured in the task-specific network (ADI) in the (task-layers) upper layers of the network.

Our analysis of the pretrained networks revealed several key findings:
i) channel and gender information are encoded and distributed throughout the networks at different layers,
ii) gender information is redundantly encoded in both the ADI and SRE pretrained models,
iii) the ADI network shows a higher susceptibility to gender-based representation bias compared to other models,
iv) all pretrained models, except for SRE, learn speaker-invariant representations,
v) voice identity is only encoded in the upper layers of the SRE model,
vi) language properties are captured in the upper layers of all pretrained models,
vii) pretrained models trained for dialect identification are well-suited for knowledge transfer in language identification downstream tasks,
viii) regional dialectal information is only captured in the task-specific network (ADI) in the upper layers of the network (task-layers), unlike language properties.

\section{Neuron Analysis}
\label{sec:resultlc}

% \sout{Next, we carried out a fine-grained neuron-level analysis for deeper insights.  We first evaluate the efficacy of the neuron ranking algorithm in Section \ref{ssec:ranking_eval}; second we compare our oracle (ALL) results with the control tasks for further validation (in Section \ref{ssec:ns1} ); and third we investigate the following questions: RQ3: what are the minimal set of neurons that capture a property (in Section \ref{ssec:ns2}), and RQ4: highlight how localized/distributed this information is (Section \ref{ssec:ns3}).}

In this section, we conduct a detailed analysis of representations at the neuron level. The section is divided into two parts: i) evaluating the effectiveness of the neuron ranking algorithm, and ii) drawing task-specific and architectural comparisons based on the identified neurons.

\begin{table}[!ht]
\centering

\scalebox{0.7}{

\begin{tabular}{|l|c|c|c|c|} 
\hline
\multicolumn{1}{|l|}{} & ADI & SRE & $ST_{base}$ & $ST_{large}$ \\ 
\hline\hline
$Neurons$ & 20\%  & 20\%  & 20\%  & 20\%  \\\hline
\hline\hline
\multicolumn{5}{|c|}{T1: GC | labels\# 2} \\ 
\hline\hline
 & \multicolumn{1}{c|}{ADI} & \multicolumn{1}{c|}{SRE} & \multicolumn{1}{c|}{$ST_{base}$} & \multicolumn{1}{c|}{$ST_{large}$} \\ 
\hline\hline

$Acc_t$ (Masked) & 98.23 & 97.58 & 95.43 & 93.65 \\\hline
$Acc_b$ (Masked) & 53.36 &  58.65 & 77.13  & 55.87 \\\hline
$Acc_r$ (Masked) & 98.06 & 87.08 & 95.93 &  92.64   \\\hline
\end{tabular}
\quad
 \begin{tabular}{|l|c|c|c|c|} 
\hline
\multicolumn{1}{|l|}{} & ADI & SRE & $ST_{base}$ & $ST_{large}$ \\ 
\hline\hline
$Neurons$ & 20\%  & 20\%  & 20\%  & 20\%  \\\hline
\hline\hline
\multicolumn{5}{|c|}{T3: LID | labels\# 7} \\ 
\hline\hline
 & \multicolumn{1}{c|}{ADI} & \multicolumn{1}{c|}{SRE} & \multicolumn{1}{c|}{$ST_{base}$} & \multicolumn{1}{c|}{$ST_{large}$} \\ 
\hline\hline
$Acc_t$ (Masked) & 65.38 & 64.89 & 50.45 & 70.51 \\ 
\hline
$Acc_b$ (Masked) & 16.30 & 17.26 & 21.13 & 17.14 \\ 
\hline
$Acc_r$ (Masked) & \multicolumn{1}{l|}{58.83} & \multicolumn{1}{l|}{42.94} & 48.40 & 58.84 \\ \hline
\end{tabular}
}
\medskip
\scalebox{0.7}{
\begin{tabular}{|l|c|c|c|c|} 
\hline
\multicolumn{5}{|c|}{T4: DID | labels\# 5} \\ 
\hline\hline
 & \multicolumn{1}{c|}{ADI} & \multicolumn{1}{c|}{SRE} & \multicolumn{1}{c|}{$ST_{base}$} & \multicolumn{1}{c|}{$ST_{large}$} \\ 
\hline\hline
$Acc_t$ (Masked) & 52.75 & 31.29 & 23.14 & 36.21 \\ 
\hline
$Acc_b $(Masked) & \multicolumn{1}{l|}{22.33} & \multicolumn{1}{l|}{23.67} & 24.21 & 22.40 \\ 
\hline
$Acc_r$ (Masked) & \multicolumn{1}{l|}{34.01} & \multicolumn{1}{l|}{25.78} & 24.97 & 29.05 \\ 
\hline
\end{tabular}
% }
\quad
% \hfil
% \scalebox{0.8}{

\begin{tabular}{|l|c|c|c|c|} 
\hline
 \multicolumn{5}{|c|}{T5:CC | labels\# 3} \\ 
\hline\hline
 & \multicolumn{1}{c|}{ADI} & \multicolumn{1}{c|}{SRE} & \multicolumn{1}{c|}{$ST_{base}$} & \multicolumn{1}{c|}{$ST_{large}$} \\ 
\hline\hline
$Acc_t$ (Masked) & 88.60 & 86.48 & 79.40 & 88.69 \\ 
\hline
$Acc_b$ (Masked) & 41.00 & 36.64 & 49.69 & 33.14 \\ 
\hline
$Acc_r$ (Masked) & 87.49 & 75.36 & 80.28 & 88.87 \\ 
\hline

\end{tabular}
}
\caption{Reported accuracy (Acc), to indicate neuron ranking and selection algorithm efficacy for the proxy Tasks T1: GC, T3:LID, T4:DID, and T5:CC using Masked 20\% t/b/r neurons. $Acc_{*}$ with \textbf{t=\textit{top}}, \textbf{b=\textit{bottom}} and \textbf{r=\textit{random}}. The rankings are based on the absolute values of the weights. | labels\# represent the number of labels in the task.
Reported performances are averaged over 5 runs. }
\label{tab:task-efficacy}
\end{table}

\subsection{Efficacy of the Neuron Ranking} 
\label{ssec:ranking_eval}

First, we evaluate the effectiveness of the neuron selection algorithm by recomputing the classifier accuracy after masking out 80\% of the neurons and retaining the top/random/bottom 20\% neurons. Please refer to Tables \ref{tab:task-efficacy} for results on different tasks (T1,3-5). Our findings reveal that the top neurons consistently yield higher accuracy compared to the bottom neurons, demonstrating the efficacy of the ranking algorithm.

% We also found this to be true comparing top versus random neurons, although in some cases, (e.g. T2:SV-$ST_{large}$), EER\footnote{Lower value indicates better performance.} the EER of the random set (15.06) is less than the EER of the top set (15.62). This results indicates that the information is redundant and distributed across the network for the studied complex tasks. We will discuss this more (in Section \ref{ssec:ns2}-Section \ref{ssec:ns3}) as we move on to our fine-grained neuron analysis.

%\subsection{Minimal Set and Task-specific Redundancy } 
\subsection{Minimal Neuron Set} 
\label{ssec:ns2}

%We further study the possibility of extracting a minimal neuron subset with comparable performance.

To analyze individual neurons, we extract a minimal subset of neurons for each task. We rank neurons using the algorithm described in Section \ref{ssec:fg-neuron} with the goal of achieving comparable performance to the oracle (within a certain threshold). Table \ref{tab:minimal-neurons} presents the minimal number of neurons extracted for each task. We discuss the results for different tasks below and also utilize the minimal set of neurons for redundancy analysis (Section \ref{ssec:redundancy}). Specifically, we define task-specific redundancy as follows: i) if the minimal set of neurons achieves at least 97\% of the oracle performance, then the remaining neurons are considered redundant for this task, ii) if randomly selected N\% of neurons also achieve the same accuracy as the top N\% neurons, we consider the information redundant for this task.

% For analyzing individual neurons, we extract a minimal subset of neurons for each task. We extract a ranking of neurons towards each auxiliary task using the algorithm described in Section \ref{ssec:fg-neuron}. We then iteratively select top N\% neurons from the ranking until it gives comparable performance to oracle (within a certain threshold). Table \ref{tab:minimal-neurons} give the minimal number of neurons extracted for each task. We discuss the results for different tasks below. We also use the minimal set of neurons to also carry out our redundancy analysis (Section \ref{ssec:redundancy}). More specifically we define task-specific redundancy as follows: i) if the minimal set of neurons achieve at least 97\% of the oracle performance then the remaining neurons are redundant w.r.t this task, ii) if randomly selected N\% neurons also achieve the same accuracy as the top N\% neurons, we say the information is redundant w.r.t this task.

%For this, we retrain the auxiliary classifier with different percentage of the selected top neurons and select the subset, performing with in the 1\% of the oracle threshold (as mentioned in Section \ref{t1layer}).

%Furthermore, using the minimal set, we study the presence of task-specific redundancy. We defined that the rest of the neurons are redundant only if (i) the accuracy of the minimal set achieve at least 97\% of the oracle performance and (ii) the accuracy of the random set is with in a threshold of 5\%.

\begin{table}[!ht]
\centering

\scalebox{0.7}{

\begin{tabular}{|l|c|c|c|c|} 
\hline
% \multicolumn{1}{|l|}{} & ADI & SRE & $ST_{base}$ & $ST_{large}$ \\ 
% \hline\hline
\multicolumn{5}{|c|}{T1: GC | labels\# 2} \\ 
\hline\hline
 & ADI & SRE & $ST_{base}$ & $ST_{large}$ \\ 
\hline\hline
$Acc$ (ALL) & 98.20 & 96.79 & 99.16 & 98.14 \\\hline
$Neu_t$ & 5\% & 50\% & 15\% & 10\% \\\hline
$Acc_t$ (Re-trained) & 98.68 & 96.54 & 98.32 & 98.14 \\\hline
$Acc_r$ (Re-trained) & 94.65 & 95.28 & 97.44 &  87.99   \\\hline
\end{tabular}
\quad
 \begin{tabular}{|l|c|c|c|c|} 
\hline
\multicolumn{5}{|c|}{T3: LID | labels\# 7} \\ 
\hline\hline
 & ADI & SRE & $ST_{base}$ & $ST_{large}$ \\ 
\hline\hline
$Acc$ (All) & 86.00 & 76.01 & 57.35 & 76.24 \\ \hline
$Neu_t$ & 20\% & 10\% & 75\% & 50\% \\ 
\hline
$Acc_t$  (Re-trained) & 85.30 & 78.97 & 57.45 & 76.43 \\ 
\hline
$Acc_r$ (Re-trained) & 82.46  & 70.00 & 55.68 & 72.53     \\\hline

\end{tabular}
}
\medskip
\scalebox{0.7}{
\begin{tabular}{|l|c|c|c|c|} 
\hline
\multicolumn{5}{|c|}{T4: DID | labels\# 5} \\ 
\hline\hline
 & ADI & SRE & $ST_{base}$ & $ST_{large}$ \\ 
\hline\hline
$Acc$ (ALL) & 55.63 & 39.12 & 36.66 & 39.22 \\ \hline
$Neu_t$ & 25\% & 5\% & 50\% & 15\% \\ 
\hline
$Acc_t$(Re-trained) & 55.43 & 40.82 & 36.01 & 38.06 \\ \hline
$Acc_r$ (Re-trained) & 50.32 & 37.52  & 33.45 & 31.45     \\\hline
\end{tabular}
% }
\quad
% \hfil
% \scalebox{0.8}{

\begin{tabular}{|l|c|c|c|c|} 
\hline
 \multicolumn{5}{|c|}{T5:CC | labels\# 3} \\ 
\hline\hline
 & ADI & SRE & $ST_{base}$ & $ST_{large}$ \\ 
\hline\hline
$Acc$ (ALL) & 93.93 & 85.51 & 86.80 & 96.55 \\ 

\hline
$Neu_t$ & 10\% & 1\% & 20\% & 10\% \\ 
\hline
$Acc_t$ (Re-trained) & 94.56 & 85.04 & 86.27 & 95.71 \\ 
\hline
$Acc_r$ (Re-trained) & 92.82 & 75.85  & 84.70  & 92.49  \\ 
\hline
\end{tabular}
}
\caption{Reported re-trained accuracy (Acc) with minimal neuron set for proxy Tasks T1,T3,T4 and T5. $Acc_{*}$ with t=\textit{top}, and r=\textit{random} \textit{Neu}: neurons. | labels\# represent the number of labels in the task. Reported performances averaged over 5 runs.  }
\label{tab:minimal-neurons}
\end{table}

\subsubsection*{T1: Gender Classification (GC)}

We have observed that a small set of neurons (5-15\%) is sufficient to achieve accuracy close to the oracle performance ($Acc (ALL)$) with an accuracy difference of within 97\%. Furthermore, in most of the pretrained models, we have observed a small accuracy difference (within a threshold of 5\%) between the top and random subsets. This indicates that the information of gender is redundantly available throughout the network.

%a presence of redundancy for the gender information throughout the network. 
%However, 
% In contrast to the rest of the pretrained models, 

% In the SRE model, we noticed that the accuracy of the probe drops when re-trained with top 50\% neurons (with respect to the masked, in Table \ref{tab:task-efficacy}, and oracle ($Acc $(ALL)) accuracy). We speculate that this behavior is due to the nature of the pretrained model and its training objective. Note that the primary objective of the pretrained SRE model is to discriminate speakers, where gender recognition is a first-line information for such discrimination. Therefore, the oracle gender classification probe, trained with all the neurons of the pretrained network, outperforms the  probe with minimal neurons,  indicating that gender-property is not redundant in the SRE model and  the neurons capture some variant information. Our hypothesis is validated when comparing the cardinality of the minimal neuron set of the SRE model ($50\%$ neurons) {\em vs} the rest pretrained models (5-15\% neurons).

In the SRE model, we have observed a drop in probe accuracy when retrained with the top 50\% neurons, as compared to the masked accuracy and the oracle (`ALL') accuracy, as shown in Table \ref{tab:task-efficacy}. We speculate that this behavior is influenced by the nature of the pretrained model and its training objective. The primary objective of the pretrained SRE model is to discriminate speakers, where gender recognition is crucial information for such discrimination. Therefore, the gender classification probe trained with all the neurons of the pretrained network outperforms the probe with minimal neurons, indicating that the gender property is not redundant in the SRE model, and the neurons capture some variant information. This hypothesis is supported by the comparison of the cardinality of the minimal neuron set in the SRE model (50\% neurons) versus the other pretrained models (5-15\% neurons).

\begin{table} [!ht]
\centering
\scalebox{0.75}{
\begin{tabular}{l|cccccc} 
\hline
EER & \multicolumn{1}{c}{$L_{b}$} & $EER(L_{b})$ & \multicolumn{1}{c}{$Neu_{t}$} & \multicolumn{1}{c}{$EER_t$} & \multicolumn{1}{c}{$EER_r$} & \multicolumn{1}{c}{$EER_b$} \\ 

\hline\hline
\multicolumn{7}{c}{EN} \\ 
\hline\hline
ADI & FC1 &22.27 & 75\% & 22.03 & 22.32 & 22.50 \\ 
\hline
SRE & FC2 & 6.81 & 75\% & 6.96 & 6.96 & 7.05 \\ 
\hline
$ST_{base}$ & L3 & 28.12 & 5\% & 27.57 & 31.04 & 32.43 \\ 
\hline
$ST_{large}$ & L11 & 32.31 & 5\% & 26.64 & 34.11 & 39.36 \\ 
\hline
\hline
\multicolumn{7}{c}{ZH} \\ 
\hline\hline
%  & \multicolumn{1}{c}{L_{b}} & Acc(L_{b}) & \multicolumn{1}{c}{Neu_{t}} & \multicolumn{1}{c}{Acc_t} & \multicolumn{1}{c}{Acc_r} & \multicolumn{1}{c}{Acc_b} \\ 
% \hline
ADI & FC1 & 13.55 & 50\% & 14.37 & 15.85 & 14.51 \\ 
\hline
SRE & FC2 & 5.47 & 50\% & 6.06 & 6.56 & 6.10 \\ 
\hline
$ST_{base}$ & L3 & 13.90 & 20\% & 13.78 & 15.19 & 20.49 \\ 
\hline
$ST_{large}$ & L11 & 15.90 & 5\% & 15.62 & 15.06 & 26.29 \\ 
\hline
\hline
\multicolumn{7}{c}{RU} \\ 
\hline\hline
%  & \multicolumn{1}{c}{L_{b}} & Acc(L_{b}) & \multicolumn{1}{c}{Neu_{t}} & \multicolumn{1}{c}{Acc_t} & \multicolumn{1}{c}{Acc_r} & \multicolumn{1}{c}{Acc_b} \\ 
% \hline
ADI & FC1 & 13.47 & 50\% & 12.81 & 14.09 & 14.66 \\ 
\hline
SRE & FC2 & 4.05 & 50\% & 4.59 & 4.41 & 4.93 \\ 
\hline
$ST_{base}$ & L3 &16.63 & 10\% & 16.25 & 16.75 & 19.29 \\ 
\hline
$ST_{large}$ & L11 & 16.27 & 10\% & 9.09 & 15.04 & 24.44 \\
\hline
\end{tabular}
}
\caption{Reported equal error rate (EER) for proxy Task T2:SV using fine-grained neuron analysis. $EER_{*}$ with t=\textit{top}, b=\textit{bottom} and r=\textit{random} \textit{Neu}: neurons. $L_b$ represent the best layer from the pretrained model with lowest EER. $Neu_t$ represent percentage of neurons selected. Reported performance averaged over 5 runs. }
\label{tab:sv_neuron}
\end{table}

\subsubsection*{T2: Speaker Verification (SV)}

In on our layer-wise results, we have observed that the speaker-variant information is only present in the last layer of the speaker recognition (SRE) model. Upon further studying the represented layer ($L_b$), we have noticed that approximately 75\% (EN) of the neurons in the last layer are utilized to represent this information, as shown in Table \ref{tab:minimal-neurons}.

When comparing top neurons to random neurons in SRE, we found that the equal error rate (EER) of the random set (6.96) is the same as the EER of the top set (6.96). However, the minimal representation obtained from the top set does not outperform the complete FC2 neuron set. This result indicates that all neurons in FC2 are relevant to the tasks.

%We will discuss this more (in Section \ref{ssec:ns2}-Section \ref{ssec:ns3}) as we move on to our fine-grained neuron analysis.

%This indicates all the neurons of the SRE model is relevant to the task.

\subsubsection*{T3: Language Identification (LID)}

In contrast to the pretrained speech transformers, the CNN models require a small neuron set (10-20\% of the total network) to encode the language property, as shown in Table \ref{tab:minimal-neurons}. Additionally, we observed that approximately 80\% of neurons are redundant for the language property only in the ADI model. We hypothesize that this redundancy is due to the nature of the task and the training objective of the model. Since the core task of ADI is to discriminate dialects, a small number of pretrained model neurons are sufficient to store the knowledge related to the language property.

\subsubsection*{T4: Regional Dialect Identification (DID)}

The layer-level analysis revealed that the ADI model captures dialectal information only in the representations learned within its network. This observation was further confirmed in the neuron analysis. Table \ref{tab:minimal-neurons} demonstrates that using only 25\% of the neurons from the ADI network, the probe achieves a comparable accuracy to the 'ALL' neuron set ($Acc (ALL)$). However, it's important to note that re-training the classifier with a random 25\% of neurons did not achieve the oracle accuracy within the accepted threshold. This suggests that, unlike other properties, the dialectal information is not redundant in the ADI network.\footnote{A huge drop in performance is also observed when experimented with top {\em vs} random Masked-Accuracy present in Table \ref{tab:task-efficacy}, reconfirming our finding.}

%Moreover we noticed that re-training with random 25\% of neurons the accuracy difference (with top neurons) is higher than the tolerated threshold. This indicates that dialectal information %are 
%is not be redundant in the ADI network.\footnote{A significant drop in performance is also noticed when experimented with top {\em vs} random Masked-Accuracy present in Table \ref{tab:task-efficacy}, thus reconfirming our finding.}

\begin{table}[!htb]
\centering
\scalebox{0.7}{
\begin{tabular}{l|cccccc}
\hline
Layers & CNN1 & CNN2 & CNN3 & CNN4 & FC1 & FC2 \\ \hline
\#Neu/Layers & 2000 & 2000 & 2000 & 3000 & 1500 & 600 \\ \hline
\#Neu (total) & \multicolumn{6}{c}{11100} \\
\hline
\hline
Tasks ($Neu_t$ \%) & \multicolumn{6}{c}{Network: ADI} \\
\hline
\hline
LID (20) & 0 (0.0\%) & 153 (1.4\%) & 0.4 (0.0\%) & 576.6 (5.2\%) & 1034.2 (9.3\%) & 455.8 (4.1\%) \\ \hline
DID (25) & 0 (0.0\%) & 162.2 (1.5\%) & 0.2 (0.0\%) & 776.8 (7.0\%) & 1264.2 (11.4\%) & 571.6 (5.1\%) \\\hline
GC (5) & 0 (0.0\%) & 10.6 (0.1\%) & 1 (0.0\%) & 74.8 (0.7\%) & 329.8 (3.0\%) & 138.8 (1.3\%) \\\hline
CC (10) & 0.8 (0.0\%) & 63.6 (0.6\%) & 4 (0.0\%) & 184 (1.7\%) & 608.8 (5.5\%) & 248.8 (2.2\%) \\
\hline
\hline
Tasks ($Neu_t$ \%) & \multicolumn{6}{c}{Network: SRE} \\
\hline
\hline
LID (10) & 25.8 (0.2\%) & 104.2 (0.9\%) & 102 (0.9\%) & 36.8 (0.3\%) & 577.2 (5.2\%) & 264 (2.4\%) \\
DID (5) & 28.8 (0.3\%) & 47.8 (0.4\%) & 56 (0.5\%) & 35.8 (0.3\%) & 382.2 (3.4\%) & 4.4 (0.0\%) \\
GC (50) & 938.2 (8.5\%) & 948.8 (8.5\%) & 936.4 (8.4\%) & 1394.8 (12.6\%) & 838 (7.5\%) & 493.8 (4.4\%) \\
CC (1) & 0 (0.0\%) & 0.4 (0.0\%) & 0 (0.0\%) & 0.6 (0.0\%) & 69.4 (0.6\%) & 40.6 (0.4\%) \\\hline
\end{tabular}
}
\caption{Distribution of top neurons across the ADI and SRE network for each task.  Reported number of neurons averaged over 5 runs. }
\label{tab:sup_dist}
\end{table}

\subsubsection*{T5: Channel Classification (CC)}

We observed that a small percentage (1-20\%) of neurons are capable of representing the property while achieving accuracy comparable to $Acc$(ALL), as shown in Table \ref{tab:minimal-neurons}. This finding highlights the pervasive nature of the channel information. Furthermore, our analysis of re-trained top and random accuracy revealed that a substantial number of neurons are redundant for representing channel information in most of the pretrained network.

%\subsection{Localised {\em vs} Distributive Nature}
\subsection{Localization {\em vs} Distributivity}
\label{ssec:ns3}

Now we highlight the parts of the networks that predominantly capture the top neurons properties. We present the distribution of the top salient (minimal) neurons across the network in Table \ref{tab:sup_dist}-\ref{tab:large_dist}, to study how distributed or localized the spread of information is.

\begin{table}
\centering
\scalebox{0.7}{
\begin{tabular}{l|ccc}
\hline
\multicolumn{1}{l}{} & \multicolumn{3}{c}{Network: $ST_{base}$} \\ 
\hline\hline
Tasks ($Neu_t$ \%) & L1 & L2 & L3 \\ 
\hline\hline
LID (75) & 605 (26.3\%) & 556.6 (24.2\%) & 566.4 (24.6\%)~ \\\hline\hline
DID (50) & 273.2 (11.9\%) & 582.4 (25.3\%) & 296.4 (12.9\%)~ \\\hline\hline
GC (15) & 42.4 (1.8\%) & 137.2 (6.0\%) & 165.4 (7.2\%)~ \\\hline\hline
CC (20) & 124.2 (5.4\%) & 231 (10.0\%) & 104.8 (4.5\%)\\\hline
\end{tabular}
}
\caption{Distribution of top neurons across the  $ST_{base}$ network for each task. Number of neurons in each layer = 768. Total neurons (`ALL') in the network = 2304. The table reports the number of neurons (\#) in each layer which is a member of $Neu_t$\%. Each cell also reports the \% the selected neurons represents wrt to the `ALL' neurons in the network. Reported number of neurons averaged over 5 runs. }
\label{tab:base_dist}
\end{table}

\subsubsection*{T1: Gender Classification (GC)}

Tables \ref{tab:sup_dist}-\ref{tab:large_dist} show that the salient neurons (e.g., 5\% of ADI network) for the gender property are predominantly present in the upper layers of the network. This finding is consistent with our previous analysis of task-specific layer- and neuron-level minimal sets, which indicates that the information is redundantly distributed. It suggests that using a small set of neurons from any part of the network (as shown in Table \ref{tab:minimal-neurons} Acc$_r$) can yield results that are close to using the entire network.

\subsubsection*{T2: Speaker Verification (SV)}

We observed from the minimal set analysis of the SRE pretrained model that a majority of the neurons (approximately 75\% in EN) in the last layer are needed to represent the information. This indicates that the information is distributed throughout the last layer, which aligns with our layer-wise observation. Furthermore, we also found that these salient neurons are shared with other properties, such as gender, in a similar manner as observed in Chinese and Russian datasets.

\begin{table}[!ht]
\centering
\scalebox{0.7}{
\begin{tabular}{l||c||c||c||c}
\hline
\multicolumn{5}{c}{Network: $ST_{large}$} \\ 
\hline\hline
\multirow{2}{*}{Layers} & \multicolumn{4}{c}{Tasks ($Neu_t$ \%)} \\ 
\cline{2-5}
 & LID (50) & DID (15) & GC (10) & CC (10) \\ 
\hline\hline
L1 & 423.6 (4.6\%) & 26.2 (0.3\%) & 11 (0.1\%) & 13.2 (0.1\%) \\
L2 & 171 (1.9\%) & 7.2 (0.1\%) & 6.2 (0.1\%) & 3 (0\%) \\
L3 & 124 (1.3\%) & 10.4 (0.1\%) & 2.4 (0\%) & 5.2 (0.1\%) \\
L4 & 153 (1.7\%) & 11.4 (0.1\%) & 2.6 (0\%) & 124.4 (1.3\%) \\
L5 & 247.6 (2.7\%) & 50.4 (0.5\%) & 12 (0.1\%) & 129.8 (1.4\%) \\
L6 & 266.8 (2.9\%) & 33.2 (0.4\%) & 3.2 (0\%) & 28.6 (0.3\%) \\
L7 & 305.6 (3.3\%) & 53.4 (0.6\%) & 10.8 (0.1\%) & 23 (0.2\%) \\
L8 & 327.6 (3.6\%) & 64.6 (0.7\%) & 9.8 (0.1\%) & 14.2 (0.2\%) \\
L9 & 615.6 (6.7\%) & 175.4 (1.9\%) & 74.6 (0.8\%) & 58.4 (0.6\%) \\
L10 & 662.2 (7.2\%) & 337.2 (3.7\%) & 227.6 (2.5\%) & 146 (1.6\%) \\
L11 & 686 (7.4\%) & 378.8 (4.1\%) & 284 (3.1\%) & 159 (1.7\%) \\
L12 & 625 (6.8\%) & 233.8 (2.5\%) & 276.8 (3\%) & 216.2 (2.3\%)\\\hline
\end{tabular}
}
\caption{Distribution of top neurons across the  $ST_{large}$ network for each task. Number of neurons in each layer = 768. Total neurons (`ALL') in the network = 9216. The table reports the number of neurons (\#) in each layer which is a member of $Neu_t$\%. Each cell also reports the \% the selected neurons represents with respect to the `ALL' neurons in the network. Reported number of neurons averaged over 5 runs. }
\label{tab:large_dist}
\end{table}

\subsubsection*{T3: Language Identification (LID)}
\label{lang_neuron}

For the language property, the information is more distributed in the pretrained transformers (see Table \ref{tab:base_dist}-\ref{tab:large_dist}), and localised in the upper layers for CNNs (see Table \ref{tab:sup_dist}). 
We hypothesize that this difference in behavior is due to the contrasting training objectives of the models, with CNNs trained with task-related supervision and Transformers trained with self-supervision. Both CNN pretrained models, ADI and SRE, are trained with objectives that involve language identification or discrimination as an innate criterion/feature for model prediction. As a result, the neurons in the upper layers of CNNs capture more language-representative information compared to their predecessors.

% We hypothesize that such a difference in the models' behaviour is due to their contrasting training objectives (CNN:task-related supervision {\em vs} Transformers:self-supervision). Both the CNN pretrained models are trained with an objective that is either a special case of language identification (dialect identification objective -- ADI) or language discrimination is one of the innate criterion/feature for model prediction (speaker recognition -- SRE). Hence the neurons in the upper layers of CNNs capture more language-representative information than its predecessor. 

% Whereas, the transformer architectures are trained using self-supervised approach -- reconstructing the masked input signals -- innately capturing language discrimination properties, distributed throughout all the transformer layers ($|$minimal neuron set $|=$ 50-75\%). 

% So, when the minimal neuron set is selected, the neurons from the last layers of CNNs are sufficient to capture the represented information. As for the transformers, the neurons are distributed throughout the network, each capturing some properties to represent language.

\subsubsection*{T4: Regional Dialect Identification (DID)}

Based on the minimal set analysis, we observed that only 25\% of the neurons in the ADI network are necessary to encode the regional dialects. Further analysis, as presented in Table \ref{tab:sup_dist}-\ref{tab:large_dist}, reveals that the regional dialectal information is predominantly localized in the upper layers of the ADI network.

\subsubsection*{T5: Channel Classification (CC)}

We found that salient neurons are localized in the upper layers of CNNs, as shown in Table \ref{tab:sup_dist}. On the other hand, in transformers, the information is distributed across the middle and upper layers, as evidenced by Table \ref{tab:base_dist}-\ref{tab:large_dist}.

\subsection{Summary}
\label{koneuron}

Our neuron analysis reveals several key findings:
i) it is possible to extract a minimal set of neurons that effectively represent the encoded information,
ii) complex tasks, such as voice identity, require a larger number of neurons to encode the information compared to simpler tasks, such as gender classification,
iii) network redundancy is observed for simple tasks, such as gender and channel identification, indicating that task-specific information is stored redundantly,
iv) for complex tasks, such as dialect and speaker voice verification, the information is not redundant and is only captured by the task-specific network, and
v) for most properties, the salient minimal neurons are localized in the upper layers of the pretrained models.

% Our neuron analysis shows that i) it is possible to extract a minimal set of neurons to represent the encoded information, ii) complex tasks (e.g., voice identity) require more neurons to encode the information compare to the simple tasks (e.g. gender classification), iii) we found network to store task-specific information redundantly for 
% simple tasks such as gender and channel identification, iv) for the complex task such as dialect and speaker voice verification, the information is not redundant and is only captured by the task-specific network, and v) for most of the properties, the salient minimal neurons are localised in the upper-layers of the pretrained models. 

\section{Discussion} 
% \subsection{From E2E Network perspective} 
\label{ssec:keyobservation}
\subsection{Key Observation from Layer and Neuron-level Analysis:}

% \nd{It is not clear how findings from layer analysis is resonating here? Below paragraph is just talking about neurons after you say "Combining our findings .."}

\paragraph{\textbf{Number of neurons and the task complexity}:} 
% From layer-wise analysis we noticed the simple properties such as gender and channel are omnipresent in the network. However, combining our findings from both layer- and neuron-level analysis, 
We observed that simple properties, such as gender and channel, can be captured with fewer neurons (as little as 1\% of the network) to encode the information. One possible explanation could be the nature of the tasks, as gender and channel information are salient and easily distinguishable in the acoustic signal, requiring only a few neurons for accurate classification. On the other hand, for more complex properties, like voice identity verification, a significant number of neurons are necessary to represent the variability in the signal.

% Our initial analysis (layer-wise) indicates that the information for gender and channel are distributed throughout the network. However, f

\paragraph{\textbf{Localized vs Distributive}:} We observed that the salient neurons for most tasks are concentrated in the upper layers of the pretrained models. This observation is particularly evident in the voice identity verification task of the SRE model, where nearly 50-75\% of neurons in the last layer encode such information. We hypothesize that neurons in each successive layer are more informative than their predecessors, and as we delve deeper into the network, more contextual information is captured, aiding in discriminating factors like variability in a speaker's voice.

\paragraph{\textbf{Task-specific Redundancy}:} We noticed task-specific redundancy at both the layer and neuron levels for gender and channel properties. This redundancy arises due to the distinct acoustic signature of these properties, which allows for the learning of redundant information across the network, often represented by a small number of neurons. In addition, we also observed neuron-level redundancy for the language property in the ADI model. We hypothesize that a limited number of neurons are sufficient to capture the variability between languages when transferred from the pretrained model, which is originally trained to distinguish dialects within a language family.

% \paragraph{\textbf{Polysemous Neurons}:} We noticed the salient neurons are sometimes shared among properties. For example, in the SRE pretrained model, we noticed that a subset (around 40\%) of the voice-identity neurons are shared with the gender neurons. This sharing properties of the neurons reflects on the main training objective -- i.e., to recognise speakers -- of the pretrained model, as both gender and voice variability is needed to verify speakers' identity. Hence, when creating speaker verification pairs to evaluate the performance of a speaker recognition model, we tend to select pairs from the same gender, removing the advantage of gender-based discrimination.
\paragraph{\textbf{Polysemous Neurons}:} We observed that salient neurons are sometimes shared across different properties. For instance, in the SRE pretrained model, we noticed that approximately 40\% of the voice-identity neurons are also shared with the gender neurons. This sharing of properties among neurons reflects the main training objective of the pretrained model, which is to recognize speakers, as both gender and voice variability are essential for verifying speakers' identity. Therefore, when creating speaker verification pairs to evaluate the performance of a speaker recognition model, we tend to select pairs from the same gender, to mitigate the influence of gender-based discrimination.

%bias and robustness
\paragraph{\textbf{Bias}:} Our analysis of fine-grained neurons reveals the presence of property-based representation bias in the pretrained network, shedding light on the specific parts (neurons) that encode information. For instance, through layer and neuron-level analysis, we demonstrate that the ADI pretrained model exhibits a higher susceptibility to gender representation bias compared to other pretrained models. By identifying neurons that capture gender-related information in the network, we can potentially manipulate and control the system's behavior to eliminate the bias. However, we leave further exploration of this topic for future research

\paragraph{\textbf{Robustness}:} Through our diagnostic tests, we observed that the pretrained networks are robust towards unknown speakers. This increases the reliability of predictions made by the pretrained models. Furthermore, leveraging this information, we can identify potential parts of the network that are susceptible to capturing identity information, and selectively fine-tune only those parts of the pretrained model for any future speaker-dependent downstream tasks, thereby reducing computational costs.

\subsection{Cross-architectural Comparison:}
% \paragraph{\textbf{Network Size and its Encoding Capabilities}:} Comparing pretrained models across different architectures, we observed that the classifiers trained from the feature representations of the smaller transformers give lower performance towards certain concepts that we studied. On the other hand, larger models learn richer feature representations. Note that the presence of these features is orthogonal to the performance of these models on their actual task (the loss function they are trained on). Determining whether these features actually improve performance on the intended task, for which the models were trained, would require conducting ablation studies. However, such studies are outside the scope of this paper.
\paragraph{\textbf{Network Size and its Encoding Capabilities}:} In comparing pretrained models across different architectures, we observed that classifiers trained from feature representations of smaller transformers exhibit lower performance towards certain concepts that we studied. On the other hand, larger models tend to learn richer feature representations. It is worth noting that the presence of these features is independent of the performance of these models on their actual task, as determined by the loss function they are trained on. Assessing whether these features actually improve performance on the intended task for which the models were trained falls beyond the scope of this paper.

\paragraph{\textbf{Storing Knowledge}:} Our observations reveal a tetrahedral pattern in the storage of knowledge, indicating that deeper neurons in the network tend to be more informative and store more knowledge compared to their counterparts in lower layers. Specifically, in large architectures such as CNNs and large transformers, we notice that task-oriented information is predominantly captured in the upper layers, encoding more abstract information, while vocal features are primarily captured in the lower layers. These findings corroborate previous studies that suggest the lower layers of the network function as feature extractors, while the upper layers serve as task-oriented classifiers. This observation is also consistent with the results presented in \cite{shah2021all}, where the authors propose that the representation in the lower layers resembles traditional low-level speech descriptors extracted from classical feature extraction pipelines, such as formants, mean pitch, and voice quality features. However, in the case of small transformer pretrained models, the information is more distributed throughout the network.

% This reinforces the previous findings that the lower layers of the network act as feature extractor and upper layers as task-oriented classifier. 
% As for the small transformer pretrained model, the information is more distributed.

% \paragraph{\textbf{Pretrained Architecture for Transfer Learning}:} For pretrain-fine-tune paradigm, the most popular architecture choice is transformers in the context of pretrained language modeling and speech representation models. However, in the research community, there is an abundance of trained large CNN models. In line with \cite{tay2021pretrained}, our findings suggest potential in (re-)using these large CNNs as pretrained model for transferring knowledge to another task, irrespective of their pretraining objectives. Our results show that re-using these pretrained CNNs, can give better/comparable performance, compared to the transformer models with less use of computational resources.

\paragraph{\textbf{Pretrained Architecture for Transfer Learning}:} The most popular architecture choice for the pretrain-fine-tune paradigm is transformers, especially in the context of pretrained language modeling and speech representation models. However, within the research community, there is also an abundance of large CNN models that have been trained for various tasks. In line with the findings of \cite{tay2021pretrained}, our results suggest that there is potential in (re-)using these large CNNs as pretrained models for transferring knowledge to another task, regardless of their original pretraining objectives. Specifically, our findings indicate that reusing these pretrained CNNs can yield comparable or even better performance compared to transformer models, with the added benefit of reduced computational resources

\subsection{Potential Applications}
\label{sec:application}
% just a briefly repeated applications
% Findings from the obtained minimal neuron set that performs comparably with the oracle (`ALL') set, for each individual studied property. This observation aligns with the findings in \cite{han2015learning}, demonstrating that fewer parameters can be used to represent the learned function of the network.

There are numerous potential applications for interpreting speech models beyond analyzing and understanding these models.  In this section, we will discuss some of these potential applications that can benefit from the findings of this work.

\paragraph{Network Pruning}  Neural Networks are inherently redundant due to massive over-parameterization, which results in computational inefficiencies in terms of storage and efficiency. However, numerous techniques have been successful in reducing the computational cost and memory requirements during inference, indicating that not all representations encoded by the rich architecture are necessary. Our neuron and redundancy analysis further reinforces this observation. Firstly, we showed that network layers are redundant with respect to different tasks. We also demonstrated that neurons within these layers are redundant and that a minimalist subset of neurons can be extracted to achieve close to oracle performance on any downstream task. These observations can be leveraged for network pruning, for example, by pruning redundant top-layers of the network to reduce its size. This not only makes the model smaller, but also results in faster inference by reducing the forward pass to the first layer that gives close to oracle performance. A study by \cite{sajjad2023:csl} also showed that pruning layers in the model is an effective way to reduce its size without significant loss in performance.

% Our results demonstrate that a minimal set of neurons can effectively represent the learned property-based information without compromising accuracy. Identifying such a small set of salient neurons can provide insights into the network's behavior and predictions. These salient neurons can be utilized for various purposes, such as network pruning or finding sparse networks, as indicated in \cite{frankle2018lottery,lai2021parp}. Additionally, these salient neurons can serve as important features for downstream tasks. Furthermore, identifying the salient neurons associated with each property can pinpoint sensitive parts of the network, as observed in Section \ref{t1layer} for the gender property in ADI.

% Interpreting speech models have numerous applications beyond analyzing and understanding these models. In the following we discuss some of the potential applications this work can be used for:

%\paragraph{Model Efficiency} 

\paragraph{Feature-based Transfer Learning} 
% Neural Networks are innately redundant due to massive over-parameterization. This makes the resulting models computationally expensive both in terms of storage and efficiency. 
% Numerous techniques have been successful in reducing the computational cost and memory requirements during inference, indicating that not all representations encoded by the rich architecture are necessary.
% Our neuron and redundancy analysis also reinforces this observation. Firstly we showed that network layers are redundant w.r.t different tasks. We also showed that neurons within these layers are redundant and that a minimalist subset of neurons can be  extracted to achieve close to oracle performance on any downstream task. Both these observation can be capitalized towards network pruning. For example if the top-layers of the network are redundant to the lower-layers, we can prune these to reduce the size of the network. This not only makes the model smaller but also results in a faster inference by reducing the forward pass to the first layer that gives close to oracle performance. \cite{poorBERT} showed that pruning layers in the model is an effective way to reduce its size without losing any significant performance. Along the same lines our neuron analysis can enable efficient feature learning. 
Feature learning has emerged as a viable alternative to fine-tuning based transfer learning in the field of Natural Language Processing (NLP), as shown by \cite{peters-etal-2019-tune}. Classifiers with large contextualized vectors suffer from issues such as being cumbersome to train, inefficient during inference, and sub-optimal when supervised data is limited, as highlighted by \cite{hameed}. To address these problems, selecting a minimal subset of relevant neurons from the network can be a solution. \cite{dalvi2020analyzing} combined layer- and neuron-analysis for efficient feature selection in NLP transformers trained on GLUE tasks \cite{wang-etal-2018-glue}. We believe that a similar methodology can benefit speech-based CNN and transformer models, and exploring this frontier is an exciting opportunity.

% Feature learning has shown to be a viable alternative to fine-tuning based transfer learning in NLP domain \cite{peters-etal-2019-tune}. Classifiers with large contextualized vectors are: a) cumbersome to train, b) inefficient during inference, and c) may be sub-optimal when supervised data is insufficient \cite{hameed}. Selecting a minimal subset of relevant neurons from the network alleviates this problem. \cite{dalvi2020analyzing} combined both layer- and neuron-analysis for efficient feature selection in NLP transformers trained towards GLUE tasks \cite{wang-etal-2018-glue}. We believe that speech based CNN and transformer models can benefit from a similar methodology and this is an exciting frontier to explore.

\paragraph{Model Manipulation} Identifying salient neurons in the model with respect to certain properties can potentially enable controlling the model with respect to that property. For example, \cite{bau:2019:ICLR} identified switch neurons that learned male and female gender verbs in Neural Machine Translation and controlled their activation values at inference to change the system's output.  In Section \ref{t1layer}, we successfully identified and located the sensitive components of the network that relate to the gender property in ADI. We believe that our neuron analysis has a promising potential in facilitating the editing and manipulation of information in speech models, providing a valuable tool for control and customization.

\subsection{Limitations}
\label{ssec:limitation}
\paragraph{Complexity of the Probe} From a methodological standpoint, we employed a straightforward logistic regression classifier, chosen for its simplicity in theoretical understanding and widespread use in the literature. However, recent studies \cite{conneau2018you} have suggested that a deeper classifier may be necessary to capture more subtle encoded knowledge. Linear probes play a crucial role in our approach, as we utilize the learned weights as a proxy to gauge the significance of each neuron. Further exploration of more complex probes is left as a potential avenue for future research.

\paragraph{Dependence on Supervision}
We trained our probes using pre-defined tasks and annotations, which enabled us to conduct layer-wise and fine-grained neuron analyses. One limitation of this approach is that our analysis is restricted to pre-defined properties for which annotations are available. To uncover additional information captured within the network, and to determine if machine-learned features align with human-engineered features, further unsupervised analysis is needed. Probing classifiers also have limitations, as the analysis can be biased by the constraints of the annotated data, such as sparsity, genre, etc. To validate our findings, it is crucial to conduct analysis under diverse data conditions. We plan to explore this avenue in future research.

% Another limitation of probing classifiers is that the analysis is biased by the limitations (sparsity, genre, etc) of the annotated data. It is important to conduct analysis under various data conditions to corroborate the findings. We leave this exploration for the future.

\paragraph{Connecting Interpretation with Prediction} 
Although probing methods are valuable in analyzing and identifying important information captured within the network, they do not necessarily reveal how this information is utilized by the network during prediction \cite{belinkov2019analysis}. For instance, to mitigate bias in the system's output, it is essential to identify neurons that are relevant to the property of interest and determine which of these neurons play a critical role during prediction. By combining these two pieces of information, one can effectively exert control over the system's behavior with respect to that property. This represents a challenging research frontier that we encourage researchers in speech modeling to explore.

\section{Conclusion}
\label{sec:concl}

In this study, we analyzed intermediate layers and salient neurons, in end-to-end speech CNN and transformer architectures for speaker (gender and voice identity), language (and its variants - dialects), and channel information.
We explored the architectures, using proxy classifiers, to determine what information is captured, where it is learned,  how distributed or focused their representation are, the minimum number of neurons required, and how the learning behavior varies with different pretrained models.
% We explored the architectures, using proxy classifiers, to investigate:  whether information is captured; 
% where is it learnt?; how distributed or focused their representation are? the minimal number of neurons we can use; and how the learning behaviour changes with different pretrained models? 

Our findings suggest that channel and gender information are omnipresent with redundant information, unlike voice identity and dialectal information, and require a small subset (mostly 1-20\%) of neurons to represent the information.
We observed, for complex tasks (e.g. dialect identification), the information is only captured in the task-oriented model, localised in the last layers, and can be encoded using a small salient (e.g., 25\% of the network) neuron set. These salient neurons are sometimes shared with other properties and are indicative of potential bias in the network. Furthermore, this study also suggests, in the era of pretrained models, in addition to popular `transformers', CNNs are also effective as pretrained models and should be explored further.

To the best of our knowledge, this is the first attempt to analyse layer-wise and neuron-level analysis, putting the pretrained speech models under the microscope.  
In future work, we plan to further extend the study to other available architecture, like autoencoders, with low-level information such as phoneme and grapheme and dig deeper into class-wise properties.

\bibliographystyle{elsarticle-num-names}
\bibliography{did_encode_info}

\newpage
\clearpage
\appendix
\section{}
\subsection{CNN Architecture}
\label{appen:cnn}
The input to the models is 40 coefficient MFCCs features from a spectrogram computed with a 25ms window and 10ms frame-rate from 16kHz audio. 
The architecture of the models includes four temporal convolution neural networks (1D-CNNs), followed by a global (statistical) pooling layer to aggregate the frame-level representations to utterance-level representations.\footnote{We followed a similar approach to extract utterance-level representation from the first 3 CNN layers for our study (see Figure \ref{fig:pipeline}).} 
For the CNN layers, we used filter sizes of 40$\times$5, 1000$\times$7, 1000$\times$1, 1000$\times$1 with 1-2-1-1 strides and 1000-1000-1000-1500 filters respectively.
This utterance-level representation is then passed to two fully connected layers (hidden units: 1500 and 600). We used Rectified Linear Units (ReLUs) as activation functions of the network.
The network is trained using the stochastic gradient descent (SGD) optimizer with a learning rate of 0.001. 

\subsection{Transformer Architecture}
\label{appen:transformer}
The input to the models is Mel-features, which are then transformed into high-level representations. For the transformation, the input is first downsampled to adapt to long input sequences, and then the consecutive frames are stacked into one step. 
This step reduces the number of frames used in the architecture. These input frames are then projected to a fixed dimension of 768 before passing to a sinusoidal function for encoding position.
As a result, these frames passed through multi-layer transformer encoder with multi-head self-attention for left-and-right bidirectional encoding. Each transformer encoder outputs the encoder's hidden states and has a dimension of 768. The transformers are trained for $50000$ steps with a learning rate of 0.001.

\subsection{ADI and SRE Model Performance}
The overall performance of the trained ADI model using the official MGB-5 dialect test set \cite{mgb5} are accuracy - $82.0$\% and $F_1$ - $82.7$\%.

To evaluate the SR model performance, we performed speaker verification, using the embedding from the last intermediate layer (second fully-connected layer, FC2) of the SRE model with Voxceleb1 official test verification pairs, obtaining $EER=6.81$.

\subsection{Variance in the Oracle Results}
To compensate for the randomness in the presented results, we ran each of our experiments 5 times and reported the average results. We noticed the variance in our results is very much insignificant, making the results replicable. We reported the standard variation in the accuracies presented in Table \ref{tab:task-baseline} in Table \ref{tab:task-baseline-std}.

\begin{table}[!ht]
\centering

\scalebox{0.7}{

\begin{tabular}{|l|c|c|c|c|c|}
\hline
Acc (ALL) & \# Class Labels & ADI      & SRE      & $ST_{base}$     & $ST_{large}$    \\ \hline
T1: GC    & 2               & $\pm$ 1.41E-03 & $\pm$ 6.36E-03 & $\pm$ 5.52E-04 & $\pm$ 2.76E-04 \\ \hline
T3: LID   & 7               & $\pm$ 3.07E-02 & $\pm$ 1.14E-02 & $\pm$ 4.38E-04 & $\pm$ 1.14E-02 \\ \hline
T4: DID   & 5               & $\pm$ 2.50E-02 & $\pm$ 1.12E-03 & $\pm$ 6.70E-04 & $\pm$ 3.00E-04 \\ \hline
T5: CC    & 3               & $\pm$ 7.02E-04 & $\pm$ 1.35E-03 & $\pm$ 7.02E-04 & $\pm$ 3.30E-04 \\ \hline
\end{tabular}}
\caption{Reported standard deviation for the average accuracy (over 5 runs) reported in Table \ref{tab:task-baseline} for Oracle (ALL: with all neurons). 
The results are reported for the proxy Tasks T1: GC, T3:LID, T4:DID and T5:CC. }
\label{tab:task-baseline-std}
\end{table}

\end{document}